\documentclass{article}

 \usepackage[preprint]{neurips_2026}

\usepackage[utf8]{inputenc} % allow utf-8 input
\usepackage[T1]{fontenc}    % use 8-bit T1 fonts
\usepackage{hyperref}       % hyperlinks
\usepackage{url}            % simple URL typesetting
\usepackage{booktabs}       % professional-quality tables
\usepackage{amsfonts}       % blackboard math symbols
\usepackage{nicefrac}       % compact symbols for 1/2, etc.
\usepackage{microtype}      % microtypography
\usepackage{xcolor}         % colors
\usepackage{graphicx}
\usepackage{wrapfig}
\usepackage{amsmath}
\usepackage{subcaption}
\usepackage{wrapfig}
\usepackage{array}
\usepackage[table]{xcolor}

\usepackage{booktabs}
\usepackage{multirow}
\hypersetup{hidelinks}

\usepackage{glossaries}

\newacronym{llms}{LLMs}{Large Language Models}
\newacronym{lra}{LRA}{Linear Residual Adapter}
\newacronym{knn}{KNN}{K-Nearest Neighbors}

\title{SHIFT-LLM \\ Distribution Shift Correction in Depth-Pruned LLMs}

\author{%
  Ali Bahri$^{1}$ \quad
  Hang Li$^{1}$ \quad
  Hongliang Li$^{1}$ \quad
  Zhitang Chen$^{2}$ \\
  $^{1}$Huawei Noah's Ark Lab, Canada \\
  $^{2}$Huawei Noah's Ark Lab, Hong Kong SAR, China
}

\begin{document}

\maketitle

\begin{abstract}

    Depth pruning removes entire Transformer blocks to reduce the inference cost of large language models, but disrupts the hidden-state distributions expected by downstream layers, leading to significant accuracy loss. We introduce SHIFT-LLM, a training-free post-pruning correction framework that inserts a \gls{lra} at each pruning site. Each \gls{lra} preserves the identity pathway of the original residual block and adds a lightweight affine residual correction. This correction is calibrated via closed-form least-squares regression on a small held-out set, without gradient computation, to approximate the missing residual update produced by the pruned block. Together with the preserved identity pathway, the resulting \gls{lra} output approximates the hidden state produced by the original block, thereby mitigating the distributional mismatch introduced by layer removal while avoiding the expensive attention and feed-forward computations of the removed blocks. The resulting LRAs support low-rank factorization and exact merging across consecutive pruned layers for additional compression, and combine naturally with parameter-efficient fine-tuning for further recovery beyond fine-tuning the pruned model alone. Experiments on five model families, six layer-selection criteria, and seven zero-shot benchmarks show that SHIFT-LLM consistently recovers accuracy lost to depth pruning across most configurations, achieving gains up to +15.7 points on Llama-3.1-8B-Instruct while requiring only a few hundred calibration samples and no gradient computation.

\end{abstract}

\section{Introduction}

Transformer-based \gls{llms} \citep{vaswani2017attention} have achieved remarkable success across many applications, but their growing scale imposes substantial computational costs that hinder practical deployment in resource-constrained settings. To address this challenge, model compression methods such as quantization \citep{frantar2022gptq, lin2024awq, xiao2023smoothquant}, knowledge distillation \citep{gu2023minillm, xu2024survey}, and pruning \citep{sun2023simple, ashkboos2024slicegpt} have been widely studied. Among them, structured pruning is especially attractive because it removes entire components, such as neurons or layers, enabling more direct hardware efficiency and inference speedup.

Pruning improves model efficiency by removing parameters or structural components while preserving performance as much as possible. Structured pruning \citep{chen2024streamlining, liu2025grasp, cao2025pip, ling2412slimgpt} removes entire neurons or layers rather than sparsifying individual weights \citep{ma2023llm, sun2023simple}, enabling more direct practical speedups and hardware efficiency. Recent studies further show that depth pruning can provide reliable small-batch latency gains while remaining competitive with structured width pruning \citep{kim2024shortened, men2025shortgpt}.

However, despite their practical appeal, existing depth pruning methods often suffer from significant performance degradation and require extensive fine-tuning to recover performance. In this work, we identify \emph{post-pruning distribution shift} as a key underlying cause of this degradation. 
\begin{wrapfigure}{r}{0.48\textwidth}
    \vspace{-3pt}
    \centering
    \includegraphics[width=0.48\textwidth]{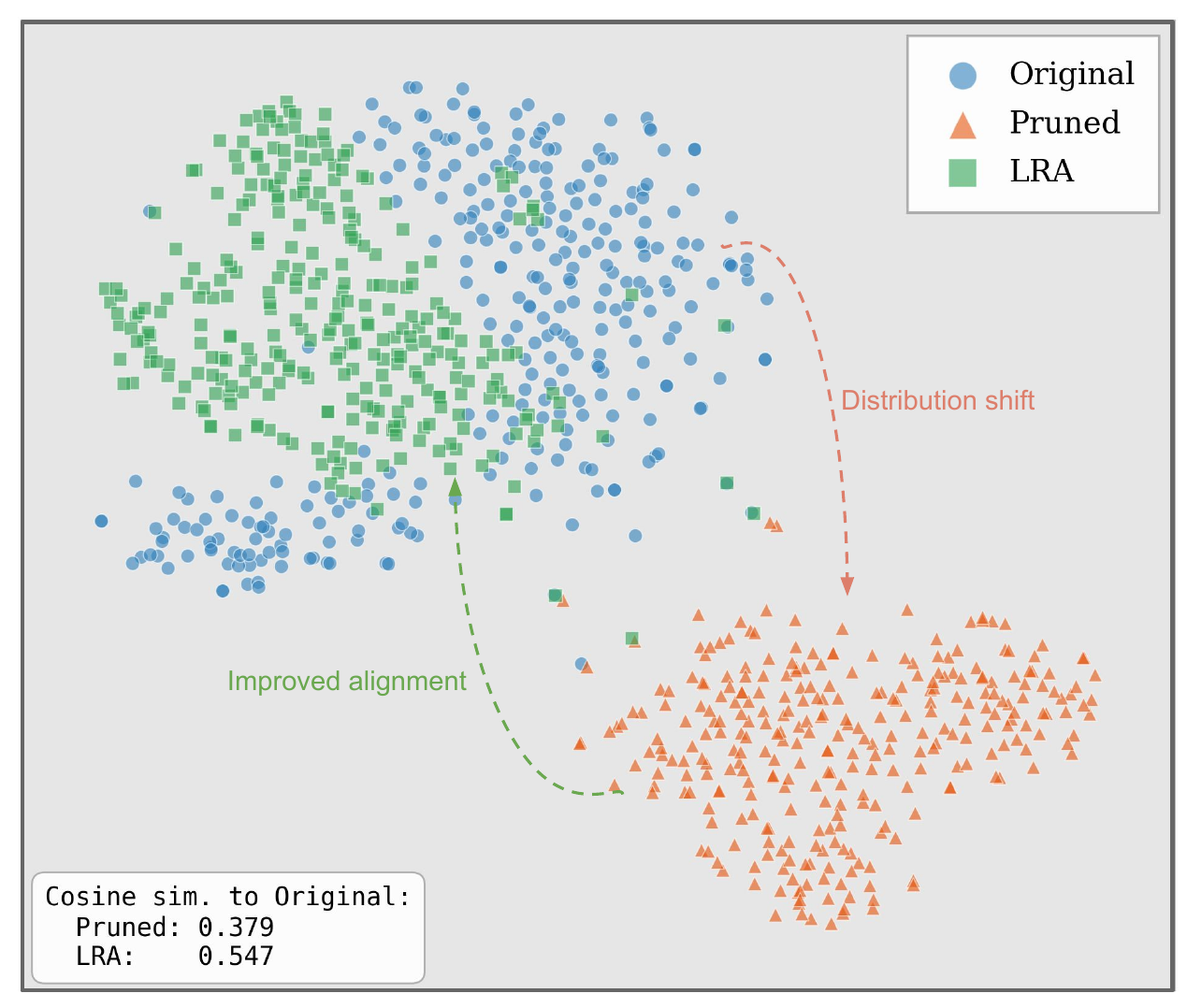}
    \caption{The proposed \gls{lra} partially restores the hidden-state distribution toward that of the original model.}
    % \vspace{-3pt}
    \label{fig:tsne_motivation}
    \vspace{-5pt}
\end{wrapfigure}
Specifically, removing an intermediate Transformer block eliminates the residual update that contributes to its output hidden state, causing downstream layers to receive hidden states that differ from those produced by the original pretrained model and therefore follow a shifted distribution. 
This shift can propagate through subsequent layers and lead to substantial performance degradation. Therefore, the challenge of depth pruning is not only to determine which blocks are removable, but also to mitigate the distribution shift induced by their removal.

% This perspective also helps explain why post-pruning fine-tuning is effective. After block removal, downstream layers receive shifted hidden-state distributions, and fine-tuning can partially adapt the remaining parameters to these altered inputs. However, this recovery is indirect, as standard fine-tuning optimizes a global task objective rather than explicitly approximating the missing residual updates introduced by pruning. This motivates a lightweight post-pruning correction mechanism that directly estimates the missing residual update, either on its own or in conjunction with fine-tuning.

This perspective also helps explain why post-pruning fine-tuning is effective. By adapting the remaining parameters to the shifted hidden-state distributions, fine-tuning can provide strong recovery, but requires additional optimization, compute, and training time. This motivates a lightweight alternative for settings where such post-pruning training is undesirable: directly approximating the missing residual update without gradient-based optimization. Moreover, when fine-tuning is available, this explicit correction can be used jointly with it to provide complementary recovery.

Motivated by this observation, we introduce SHIFT-LLM, a \emph{training-free} post-pruning correction framework for depth-pruned LLMs. Its core component is the \acrfull{lra}, which preserves the identity pathway of the original residual block and adds a lightweight affine residual correction. This correction is calibrated via closed-form least-squares regression on a small held-out set, without gradient computation, to approximate the missing residual update. Together with the preserved identity pathway, the \gls{lra} output approximates the hidden state produced by the original block, thereby mitigating the post-pruning distribution shift while avoiding its expensive attention and feed-forward computations.

This formulation relies on two key design principles. First, existing layer-selection methods are designed to prune redundant or less critical layers, making their missing residual updates more amenable to low-complexity approximation and allowing a lightweight affine correction to capture their dominant effect. 
Second, the \gls{lra} preserves the residual structure of the original Transformer block by retaining the identity pathway and approximating only the missing residual update. This makes the approximation problem easier than directly reconstructing the full output hidden state, since the affine residual correction only needs to approximate the missing residual update. Moreover, retaining the identity pathway preserves the residual computation structure on which the original model was pretrained, allowing the corrected hidden state to better match the output of the original block.
Figure~\ref{fig:tsne_motivation} supports this view: the pruned model exhibits a clear shift in hidden-state space, whereas the proposed \gls{lra} moves the hidden states substantially closer to the original distribution. For additional efficiency, the resulting \gls{lra}s can be compressed via low-rank factorization and merged across consecutive pruning sites. Overall, these results suggest that approximating the missing residual updates while preserving the residual structure provides a simple, general, and effective complement to depth pruning.
Our contributions are summarized as follows:

\begin{itemize}

    \item We introduce SHIFT-LLM, a \emph{training-free} framework that explicitly corrects the distribution shift caused by structured depth pruning in \gls{llms}. Its core \acrfull{lra} preserves the identity pathway of the original residual block, while its affine residual correction approximates the missing residual update introduced by layer removal.
    
    \item SHIFT-LLM is a general and modular post-pruning correction framework that can be applied across a wide range of existing layer-selection strategies for depth pruning, without being tied to any specific pruning criterion.

    % \item We conduct extensive experiments showing that SHIFT-LLM consistently improves the average performance of depth-pruned models across diverse \gls{llms}, pruning strategies, and evaluation settings, both with and without post-pruning fine-tuning.

    \item We conduct extensive experiments showing that SHIFT-LLM provides strong training-free recovery across diverse \gls{llms}, pruning strategies, and evaluation settings, while remaining complementary to post-pruning fine-tuning and further improving its performance in most evaluated settings.

\end{itemize}

\section{Related Works}

LLM layer pruning aims to improve inference efficiency by removing selected Transformer blocks while preserving downstream performance. Existing methods typically rank layers using importance criteria based on weight statistics, gradient-based saliency, activation patterns, or the loss change caused by layer removal, and then prune the least important blocks to reduce computation and memory usage.

Several representative methods have explored this direction. LLM-Pruner~\citep{ma2023llm} performs task-agnostic structured pruning using gradient-based importance and recovers performance with lightweight tuning such as LoRA. LaCo~\citep{yang2024laco} compresses models by folding later layers into earlier ones, while SlimGPT~\citep{ling2024slimgpt} combines fast near-optimal pruning with non-uniform pruning ratios to reduce cross-layer error accumulation. Shortened LLaMA~\citep{kim2024shortened} ranks layers using magnitude, Taylor sensitivity, and perplexity-based criteria, whereas ShortGPT~\citep{men2025shortgpt} introduces Block Influence (BI), which measures layer redundancy through the similarity between layer inputs and outputs. SlimLLM~\citep{guo2025slimllm} further estimates the importance of channels and attention heads at the sub-module level and uses linear regression for efficient performance recovery.

LLM-Streamline~\citep{chen2024streamlining} prunes consecutive low-importance layers based on their effect on hidden states and uses a lightweight replacement module, together with a stability metric, to reduce performance loss. GRASP~\citep{liu2025grasp} preserves gradient-sensitive singular components and replaces redundant layers with a compact parameterization estimated from a small calibration set. ReplaceMe~\citep{shopkhoev2025replaceme} is a concurrent training-free depth pruning method that compensates for removed Transformer blocks using a linear transformation fitted on a small calibration set. The key difference lies in the correction formulation: ReplaceMe regresses through a neighboring layer to approximate the effect of the removed block, whereas SHIFT-LLM performs a local correction at each pruning site by estimating only the missing residual update and preserving the original identity pathway. A detailed theoretical and empirical comparison is provided in the supplementary material.
PIP~\citep{cao2025pip} performs structured pruning by comparing clean and perturbed input views and using gradient differences to identify less sensitive components. Navigation LLM Layer Pruning~\citep{lu2024reassessing} provides a large-scale empirical study showing that simple layer-pruning strategies combined with fine-tuning of the output head and recent layers can be highly effective. Finally, Siddiqui et al.~\citep{siddiqui2024deeper} study simple additive corrections and trained low-rank linear adapters in place of removed blocks, showing that lightweight linear modules can partially recover depth-pruning performance, albeit through gradient-based fitting rather than a closed-form training-free formulation.

\section{Method}
\label{method}

\begin{figure*}[t]
    \centering
    \includegraphics[width=\textwidth,height=\textheight,keepaspectratio]{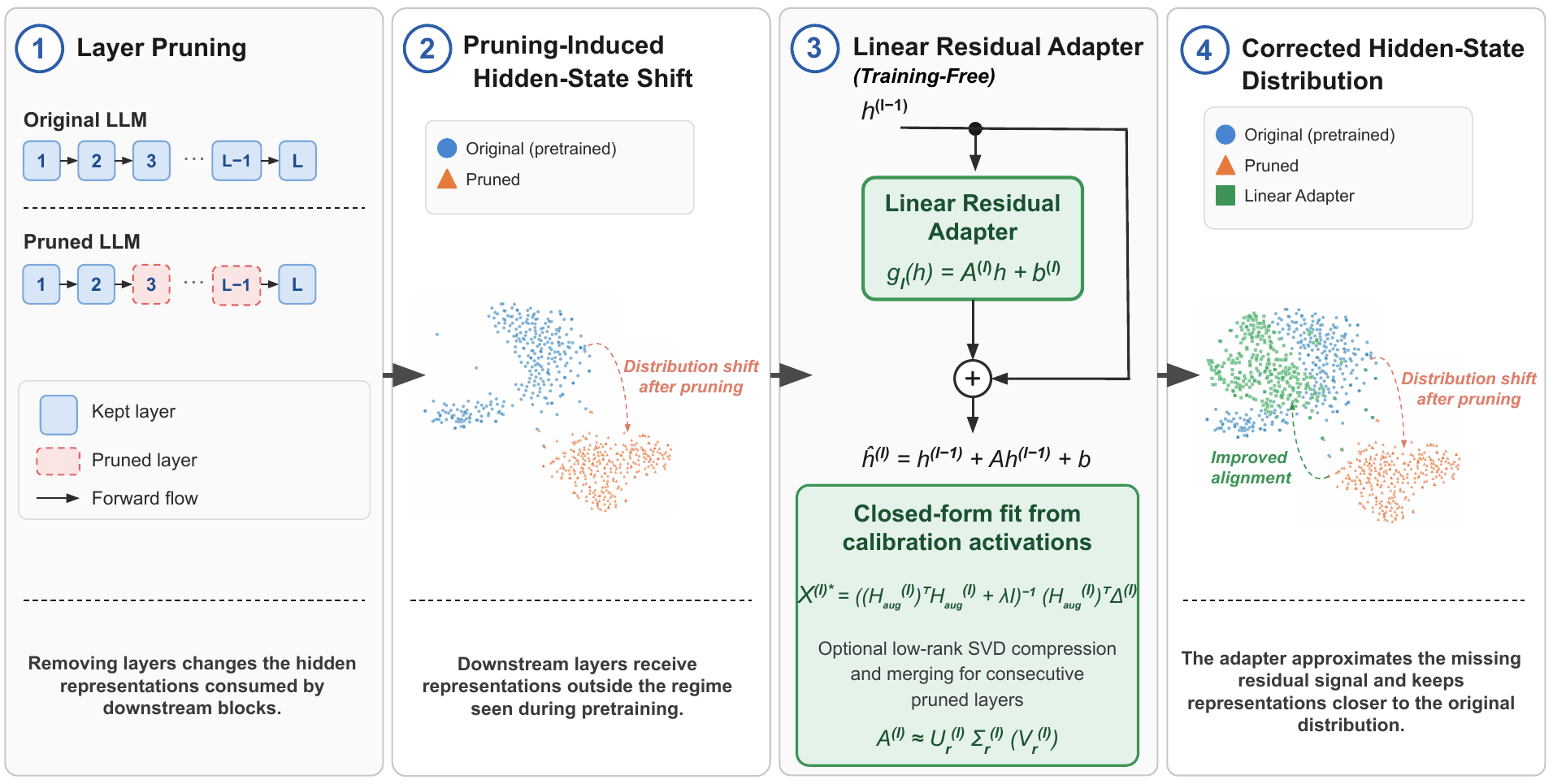}
    \caption{Overview of SHIFT-LLM for depth pruning in \gls{llms}. Depth pruning induces a hidden-state distribution shift. At each pruning site, the \gls{lra} preserves the identity pathway and adds an affine residual correction estimated in closed form from calibration activations to approximate the missing residual update. The resulting corrected hidden states remain closer to the original distribution, while the LRAs support low-rank compression and exact merging across consecutive pruning sites.}
    \label{fig:main}
\end{figure*}

In this section, we introduce SHIFT-LLM, our training-free post-pruning correction framework for mitigating the hidden-state distribution shift caused by depth pruning. We first formalize the effect of layer removal on downstream hidden states, then present its core component, the \acrfull{lra}. The \gls{lra} preserves the identity pathway of the original residual block and adds an affine residual correction that approximates the missing residual update. Together, they produce a corrected hidden state that approximates the output hidden state of the original block. We further derive the closed-form estimation procedure and discuss practical extensions, including low-rank compression and exact merging across consecutive pruning sites. Figure~\ref{fig:main} provides an overview of SHIFT-LLM.

\subsection{Pruning-Induced Hidden-State Shift}
\label{sec:hidden_shift}

We consider a Transformer composed of residual blocks, where the hidden representation at layer $\ell$ is given by
\begin{equation}
h^{(\ell)} = h^{(\ell-1)} + f_\ell\!\left(h^{(\ell-1)}\right),
\label{eq:residual_block}
\end{equation}
where $h^{(\ell-1)} \in \mathbb{R}^{T \times d}$ denotes the input hidden states, $T$ is the sequence length, $d$ is the hidden dimension, and $f_\ell(\cdot): \mathbb{R}^{T \times d} \rightarrow \mathbb{R}^{T \times d}$ denotes the transformation implemented by the $\ell$-th block, including the attention and feed-forward computations.
In standard depth pruning, removing layer $\ell$ bypasses this transformation entirely. The resulting hidden states after pruning become
\begin{equation}
\tilde{h}^{(\ell)} = h^{(\ell-1)},
\label{eq:pruned_block}
\end{equation}
instead of the original output $h^{(\ell)}$. Consequently, the subsequent block receives representations that differ systematically from those observed during pretraining.
The missing residual update introduced by pruning can be written as
\begin{equation}
\delta^{(\ell)}
= h^{(\ell)} - \tilde{h}^{(\ell)}
= f_\ell\!\left(h^{(\ell-1)}\right),
\qquad
\delta^{(\ell)} \in \mathbb{R}^{T \times d}.
\label{eq:missing_residual}
\end{equation}
Equation~\eqref{eq:missing_residual} shows that depth pruning removes the residual update produced by the pruned block. As a result, downstream layers receive hidden states whose distribution differs from that produced by the original pretrained model. This shift is particularly harmful in deep residual architectures because the output hidden state of each block becomes the input to subsequent blocks. When layer $\ell$ is removed, block $\ell+1$ receives $\tilde{h}^{(\ell)}$ instead of the original hidden state $h^{(\ell)}$, and the resulting shift can propagate through the remaining network.

SHIFT-LLM is motivated by this observation. Rather than viewing layer pruning solely as the removal of redundant computation, we also consider the hidden-state distribution shift induced by the missing residual update. This suggests that an effective post-pruning correction should approximate the missing residual update so that the corrected hidden state remains close to that produced by the original block, while preserving the computational benefits of depth pruning.

\subsection{Linear Residual Adapter Formulation}
\label{sec:adapter_formulation}

Motivated by the analysis above, we insert a lightweight \gls{lra} at each pruning site to mitigate the hidden-state distribution shift caused by layer removal. Suppose that layer $\ell$ is pruned from the original network. Rather than directly forwarding $h^{(\ell-1)}$ to the subsequent block, the \gls{lra} preserves this identity pathway and adds an affine residual correction $g_\ell(\cdot)$ that approximates the missing residual update. The corrected hidden state is therefore defined as

\begin{equation}
\hat{h}^{(\ell)} = h^{(\ell-1)} + g_\ell\!\left(h^{(\ell-1)}\right),
\label{eq:corrected_hidden}
\end{equation}
where $g_\ell(\cdot): \mathbb{R}^{T \times d} \rightarrow \mathbb{R}^{T \times d}$ denotes the affine residual correction used to approximate the missing residual update $\delta^{(\ell)}$.
Our objective is therefore to approximate the missing residual update $\delta^{(\ell)}$ defined in Eq.~\eqref{eq:missing_residual}. To this end, we model the affine residual correction as
\begin{equation}
g_\ell(h) = h A^{(\ell)} + \mathbf{1}(b^{(\ell)})^\top,
\label{eq:affine_adapter}
\end{equation}
where $A^{(\ell)} \in \mathbb{R}^{d \times d}$, $b^{(\ell)} \in \mathbb{R}^{d}$, and $\mathbf{1} \in \mathbb{R}^{T}$ broadcasts the bias over all token positions. Substituting Eq.~(\ref{eq:affine_adapter}) into Eq.~(\ref{eq:corrected_hidden}) yields
\begin{equation}
\hat{h}^{(\ell)}
=
h^{(\ell-1)}
+
h^{(\ell-1)} A^{(\ell)}
+
\mathbf{1}(b^{(\ell)})^\top
=
h^{(\ell-1)}\left(I + A^{(\ell)}\right)
+
\mathbf{1}(b^{(\ell)})^\top.
\label{eq:corrected_affine}
\end{equation}

The affine residual correction $g_\ell(\cdot)$ provides a local approximation of the missing residual update in Eq.~\eqref{eq:missing_residual}. Importantly, the \gls{lra} does not reconstruct the entire output hidden state from scratch: it preserves the identity pathway and estimates only the contribution removed by pruning. This simplifies the approximation problem while retaining the residual computation structure on which the original model was pretrained. Together, the identity pathway and affine residual correction produce a corrected hidden state that approximates the output of the original block, thereby mitigating the pruning-induced distribution shift.

The \gls{lra} is designed to satisfy three practical requirements. First, it should be substantially cheaper than the original Transformer block so that the efficiency benefits of depth pruning are preserved. Second, its affine residual correction should be estimable from a small calibration set without gradient-based optimization. Third, it should operate locally at each pruning site without modifying the remaining network.

With this formulation, the pruned model no longer propagates the uncorrected hidden state $\tilde{h}^{(\ell)} = h^{(\ell-1)}$, but instead uses the corrected hidden state in Eq.~\eqref{eq:corrected_affine}. Since each \gls{lra} defines an affine mapping through the preserved identity pathway and affine residual correction, it naturally supports low-rank factorization and exact merging across consecutive pruning sites, as discussed in Sec.~\ref{sec:extensions_complexity}.

\subsection{Closed-Form Estimation}
\label{sec:closed_form_estimation}

The affine residual correction of the \gls{lra} is estimated from a small calibration set without gradient-based optimization. The original unpruned model is run once on the calibration samples, and for each pruned layer $\ell$, the input hidden states and corresponding residual updates are collected via forward hooks. Stacking the token-level inputs yields $H^{(\ell)} \in \mathbb{R}^{N \times d}$, where each row corresponds to a sampled token representation from $h^{(\ell-1)}$, and $N$ denotes the total number of collected tokens across all calibration samples. For the same samples, we stack the missing residual updates defined in Eq.~\eqref{eq:missing_residual} into $\Delta^{(\ell)} \in \mathbb{R}^{N \times d}$, with
$\Delta^{(\ell)}_{n,:} = h^{(\ell)}_{n,:} - h^{(\ell-1)}_{n,:}$.

Using the affine residual correction defined in Eq.~\eqref{eq:affine_adapter}, our objective is to estimate $A^{(\ell)}$ and $b^{(\ell)}$ such that $g_\ell(h)$ approximates the missing residual update. To jointly estimate the weight matrix and bias term, we augment the input matrix with a column of ones:
\begin{equation}
H_{\mathrm{aug}}^{(\ell)} = \left[ H^{(\ell)} \mid \mathbf{1} \right] \in \mathbb{R}^{N \times (d+1)},
\label{eq:H_aug}
\end{equation}
and define the stacked parameter matrix
\begin{equation}
X^{(\ell)} =
\begin{bmatrix}
A^{(\ell)} \\
(b^{(\ell)})^\top
\end{bmatrix}
\in \mathbb{R}^{(d+1)\times d}.
\label{eq:X_def}
\end{equation}
The parameters of the affine residual correction are then obtained by solving the ridge-regression problem
\begin{equation}
X^{(\ell)\,*}
=
\arg\min_{X}
\left\|
H_{\mathrm{aug}}^{(\ell)} X - \Delta^{(\ell)}
\right\|_F^2
+
\lambda
\left\| X \right\|_F^2,
\label{eq:ridge_objective}
\end{equation}
where $\lambda > 0$ is a regularization coefficient.
The minimizer of Eq.~(\ref{eq:ridge_objective}) has the closed-form solution
\begin{equation}
X^{(\ell)\,*}
=
\left(
(H_{\mathrm{aug}}^{(\ell)})^\top H_{\mathrm{aug}}^{(\ell)}
+
\lambda I
\right)^{-1}
(H_{\mathrm{aug}}^{(\ell)})^\top
\Delta^{(\ell)}.
\label{eq:closed_form_solution}
\end{equation}
The first $d$ rows of $X^{(\ell)\,*}$ define $A^{(\ell)}$, and the last row defines $(b^{(\ell)})^\top$. Once estimated, the \gls{lra} is applied as
\begin{equation}
\hat{h}^{(\ell)}
=
h^{(\ell-1)}
+
h^{(\ell-1)}A^{(\ell)}
+
\mathbf{1}(b^{(\ell)})^\top.
\label{eq:final_adapter_application}
\end{equation}

This estimation procedure is entirely training-free: it requires only forward passes through the original model to collect calibration activations, followed by a single linear solve for each pruned layer. It therefore avoids iterative backpropagation, optimizer tuning, and recovery fine-tuning, while providing a closed-form estimate of the affine residual correction that approximates the missing residual update.

\subsection{Extensions and Complexity}
\label{sec:extensions_complexity}

The affine structure of the proposed \gls{lra} enables two practical efficiency extensions without changing the closed-form estimation procedure. First, the fitted weight matrix $A^{(\ell)}$ can be compressed by truncated singular value decomposition,
\begin{equation}
A^{(\ell)} \approx U_r^{(\ell)} \Sigma_r^{(\ell)} (V_r^{(\ell)})^\top,
\label{eq:svd_truncation}
\end{equation}
where $r \ll d$ is the retained rank. The affine residual correction can then be written as
\begin{equation}
g_\ell(h) \approx h P^{(\ell)} Q^{(\ell)} + \mathbf{1}(b^{(\ell)})^\top,
\label{eq:low_rank_adapter}
\end{equation}
with $P^{(\ell)} = U_r^{(\ell)}\Sigma_r^{(\ell)}$ and $Q^{(\ell)} = (V_r^{(\ell)})^\top$. This reduces the parameter count from $d^2+d$ to $2dr+d$, which is particularly useful when pruned layers are non-consecutive and require separate local \glspl{lra}.

Second, consecutive \glspl{lra} can be merged exactly. Since each \gls{lra} defines an affine mapping, it can be written as
\begin{equation}
\hat{h}^{(\ell)}
=
h^{(\ell-1)}M^{(\ell)}
+
\mathbf{1}(b^{(\ell)})^\top,
\qquad
M^{(\ell)} = I + A^{(\ell)}.
\label{eq:affine_M}
\end{equation}
Because the composition of affine mappings remains affine, consecutive \glspl{lra} can be collapsed into a single transformation without introducing additional approximation. Thus, SHIFT-LLM supports both low-rank compression and exact merging while preserving the efficiency benefits of depth pruning, as further quantified through computational cost and runtime analysis in Sec.~\ref{sec:computational_efficiency}.

\section{Experiments}
\label{Experiments}

In this section, we evaluate SHIFT-LLM across different model families, layer-selection criteria, and recovery settings. We first assess its training-free effectiveness without post-pruning fine-tuning, including zero-shot accuracy, language-modeling perplexity, and scaling to a 14B model. We then evaluate its compatibility with post-pruning fine-tuning and compare it with existing pruning pipelines. Finally, we analyze the main design choices and practical efficiency of the proposed framework. Additional experiments and implementation details are provided in the supplementary material.

\begin{table*}[t]
\centering
\caption{Zero-shot average accuracy across seven evaluation benchmarks for different pruning criteria without fine-tuning at a pruning ratio of 25\%. ``Original'' denotes the unpruned model before layer pruning. ``Base'' denotes the pruned model without post-pruning correction, while ``+LRA'' denotes the same pruned model equipped with the proposed \gls{lra}. ``Gain'' reports the average improvement obtained by adding the \gls{lra}. Full per-benchmark results for each model are provided in the supplementary material.}
\footnotesize
\setlength{\tabcolsep}{3pt}
\resizebox{0.9\textwidth}{!}{%
\begin{tabular}{lccc|ccc|ccc|ccc}
\toprule
& \multicolumn{3}{c|}{Qwen2-1.5B}
& \multicolumn{3}{c|}{Qwen1.5-7B}
& \multicolumn{3}{c|}{Llama-3.1-8B-It}
& \multicolumn{3}{c}{Vicuna-7B} \\
\cmidrule(lr){2-4} \cmidrule(lr){5-7} \cmidrule(lr){8-10} \cmidrule(lr){11-13}
Original & \multicolumn{3}{c|}{\underline{59.87}} & \multicolumn{3}{c|}{\underline{65.54}} & \multicolumn{3}{c|}{\underline{71.03}} & \multicolumn{3}{c}{\underline{66.80}} \\
\midrule
Pruning Criteria & Base & +LRA & Gain
& Base & +LRA & Gain
& Base & +LRA & Gain
& Base & +LRA & Gain \\
\midrule
Block Influence   & 45.78 & \textbf{46.85} & \textcolor{gray!100!100}{+1.08} & 52.06 & \textbf{52.82} & \textcolor{gray!100}{+0.76} & 57.18 & \textbf{61.62} & \textcolor{gray!100}{+4.44} & 55.05 & \textbf{55.52} & \textcolor{gray!100}{+0.48} \\
Reverse-order*     & 43.71 & \textbf{45.71} & \textcolor{gray!100}{+2.00} & 49.25 & \textbf{53.79} & \textcolor{gray!100}{+4.54} & 43.33 & \textbf{59.06} & \textcolor{gray!100}{+15.74} & \textbf{53.47} & 52.96 & \textcolor{gray!100}{-0.51} \\
Magnitude-$\ell_1$ & 47.44 & \textbf{48.47} & \textcolor{gray!100}{+1.03} & 40.72 & \textbf{47.59} & \textcolor{gray!100}{+6.88} & 37.81 & \textbf{39.42} & \textcolor{gray!100}{+1.61} & 34.86 & \textbf{35.09} & \textcolor{gray!100}{+0.22} \\
Magnitude-$\ell_2$ & 43.76 & \textbf{45.78} & \textcolor{gray!100}{+2.02} & 43.83 & \textbf{45.74} & \textcolor{gray!100}{+1.91} & 36.00 & \textbf{39.58} & \textcolor{gray!100}{+3.58} & 35.10 & \textbf{37.03} & \textcolor{gray!100}{+1.92} \\
Taylor            & 45.93 & \textbf{46.91} & \textcolor{gray!100}{+0.98} & 51.19 & \textbf{53.04} & \textcolor{gray!100}{+1.85} & 43.30 & \textbf{59.02} & \textcolor{gray!100}{+15.71} & 56.66 & \textbf{56.67} & \textcolor{gray!100}{+0.01} \\
PPL               & 45.73 & \textbf{47.76} & \textcolor{gray!100}{+2.03} & 35.89 & \textbf{37.53} & \textcolor{gray!100}{+1.64} & 61.17 & \textbf{62.21} & \textcolor{gray!100}{+1.05} & 55.34 & 50.92 & \textcolor{gray!100}{-4.43} \\
\bottomrule
\end{tabular}%
}
\label{tab:main_no_ft_summary}
\end{table*}

\begin{table*}[t]
\centering
\caption{Zero-shot results with fine-tuning on C4 under Reverse-order$^{*}$ pruning. ``Orig.'' denotes the unpruned model, ``LoRA'' denotes standard post-pruning LoRA fine-tuning, and ``LoRA+LRA'' denotes LoRA fine-tuning combined with the proposed \gls{lra}, whose affine residual correction is kept trainable during fine-tuning.}
\scriptsize
\setlength{\tabcolsep}{2.6pt}
\renewcommand{\arraystretch}{1.08}
% \begin{tabular}{lccc|ccc|ccc|ccc}
\begin{tabular}{l
c!{\color{gray!50}\vrule width 0.3pt}cc|
c!{\color{gray!50}\vrule width 0.3pt}cc|
c!{\color{gray!50}\vrule width 0.3pt}cc|
c!{\color{gray!50}\vrule width 0.3pt}cc}
\toprule
\multirow{2}{*}{Dataset}
& \multicolumn{3}{c|}{Qwen2-1.5B}
& \multicolumn{3}{c|}{Qwen1.5-7B}
& \multicolumn{3}{c|}{Llama-3.1-8B-It}
& \multicolumn{3}{c}{Vicuna-7B} \\
\cmidrule(lr){2-4} \cmidrule(lr){5-7} \cmidrule(lr){8-10} \cmidrule(lr){11-13}
& Orig. & LoRA & LoRA+LRA
& Orig. & LoRA & LoRA+LRA
& Orig. & LoRA & LoRA+LRA
& Orig. & LoRA & LoRA+LRA \\
\midrule
BoolQ      & 73.36 & 63.15 & \textbf{65.54} & 81.9 & 71.28 & \textbf{73.24} & 83.98 & \textbf{70.31} & 69.11 & 80.55 & 74.34 & \textbf{75.99} \\
PIQA       & 75.46 & \textbf{64.31} & 64.25 & 78.40 & 68.72 & \textbf{69.91} & 79.98 & 71.71 & \textbf{72.25} & 77.26 & 70.51 & \textbf{71.16} \\
HellaSwag  & 65.41 & 45.64 & \textbf{49.18} & 76.97 & 61.70 & \textbf{63.54} & 79.27 & 67.95 & \textbf{70.43} & 73.76 & 66.14 & \textbf{67.28} \\
WinoGrande & 66.22 & \textbf{58.41} & 57.93 & 66.14 & \textbf{59.98} & 58.09 & 74.19 & 65.98 & \textbf{66.14} & 69.61 & \textbf{65.19} & 63.38 \\
ARC-e      & 66.16 & 45.03 & \textbf{46.30} & 70.71 & \textbf{58.75} & 58.00 & 81.78 & 63.22 & \textbf{66.37} & 75.63 & 62.75 & \textbf{63.80} \\
ARC-c      & 36.09 & 28.16 & \textbf{30.38} & 42.83 & 35.32 & \textbf{38.57} & 55.03 & 43.52 & \textbf{44.71} & 45.82 & 39.76 & \textbf{40.27} \\
OBQA       & 36.40 & 28.00 & \textbf{29.00} & 41.80 & \textbf{32.80} & 32.60 & 43.00 & 37.20 & \textbf{38.80} & 45.00 & 37.00 & \textbf{38.80} \\
\midrule
Avg.       & 59.87 & 47.53 & \textbf{48.94} & 65.54 & 55.51 & \textbf{56.28} & 71.03 & 59.98 & \textbf{61.12} & 66.80 & 59.38 & \textbf{60.10} \\
\bottomrule
\end{tabular}
\label{tab:main_ft_reverse_lora}
\end{table*}

\subsection{Main Results Without Fine-Tuning}
\label{sec:main_results_no_ft}

Table~\ref{tab:main_no_ft_summary} summarizes the zero-shot performance of depth-pruned models without post-pruning fine-tuning across four \gls{llms}, averaged over seven benchmarks. Overall, adding the proposed \gls{lra} improves the pruned baselines across most pruning criteria and models. The affine residual correction approximates the missing residual update, while the preserved identity pathway helps keep the corrected hidden state closer to that produced by the original block.

The \gls{lra} improves all pruning criteria on Qwen2-1.5B and Qwen1.5-7B, with gains up to \textbf{+2.03} and \textbf{+6.88}, respectively, and achieves gains of \textbf{+15.74} and \textbf{+15.71} on Llama-3.1-8B-Instruct under Reverse-order$^{*}$ and Taylor pruning. 
Gains are smaller and occasionally negative on Vicuna-7B, suggesting that some pruning criteria may select blocks whose missing residual updates remain too complex for the token-wise affine correction. Supplementary experiments support this explanation: changing the calibration domain does not remove these negative gains, while random layer selection further degrades performance, indicating that SHIFT-LLM is most effective when the pruning criterion identifies blocks whose residual updates are sufficiently amenable to low-complexity approximation.
Overall, the results show that SHIFT-LLM provides effective training-free correction across diverse pruning criteria and model families.

\paragraph{Scaling to Qwen2.5-14B.}
We further evaluate SHIFT-LLM on Qwen2.5-14B at 25\% depth pruning. The \gls{lra} improves average zero-shot accuracy from 53.34 to 56.32 under Block Influence (\textbf{+2.98}) and from 46.72 to 52.08 under Reverse-order$^{*}$ pruning (\textbf{+5.36}). These results show that the proposed training-free correction remains effective at the 14B scale. Full per-benchmark results are provided in the supplementary material.

\begin{wraptable}{r}{0.47\columnwidth}
    \vspace{-10pt}
    \centering
    \scriptsize
    \setlength{\tabcolsep}{3.2pt}
    \renewcommand{\arraystretch}{1.02}
    \begin{tabular}{llccc}
        \toprule
        Model & Criterion & Base & +LRA & Gain \\
        \midrule
        Qwen2-1.5B & BI & 87.16 & \textbf{52.98} & \textcolor{gray!100}{+34.18} \\
        & Reverse$^{*}$ & 458.44 & \textbf{111.33} & \textcolor{gray!100}{+347.11} \\
        \arrayrulecolor{gray!30}\cmidrule(lr){1-5}\arrayrulecolor{black}
        Qwen1.5-7B & BI & 146.65 & \textbf{54.66} & \textcolor{gray!100}{+91.99} \\
        & Reverse$^{*}$ & 130.21 & \textbf{64.12} & \textcolor{gray!100}{+66.09} \\
        \arrayrulecolor{gray!30}\cmidrule(lr){1-5}\arrayrulecolor{black}
        Llama-3.1-8B & BI & 161.51 & \textbf{33.83} & \textcolor{gray!100}{+127.68} \\
        & Reverse$^{*}$ & 1179.23 & \textbf{51.37} & \textcolor{gray!100}{+1127.86} \\
        \arrayrulecolor{gray!30}\cmidrule(lr){1-5}\arrayrulecolor{black}
        Vicuna-7B & BI & 41.63 & \textbf{25.52} & \textcolor{gray!100}{+16.11} \\
        & Reverse$^{*}$ & \textbf{43.94} & 71.37 & \textcolor{gray!100}{-27.43} \\
        \bottomrule
    \end{tabular}
    \caption{WikiText-2 perplexity at 25\% depth pruning. Lower is better. Gain denotes the perplexity reduction obtained by adding the \gls{lra}.}
    \label{tab:wikitext_ppl}
    \vspace{-30pt}
\end{wraptable}

\paragraph{Language Modeling Evaluation.}
To evaluate performance beyond zero-shot classification, we measure WikiText-2 perplexity at 25\% depth pruning while using C4 only for calibration. As shown in Table~\ref{tab:wikitext_ppl}, the proposed \gls{lra} substantially reduces perplexity across both pruning criteria for Qwen2-1.5B, Qwen1.5-7B, and Llama-3.1-8B, and under BI pruning for Vicuna-7B. 
The exception is Vicuna-7B under Reverse-order$^{*}$ pruning, consistent with the weaker recovery observed in Table~\ref{tab:main_no_ft_summary}.

\subsection{Main Results With Fine-Tuning}
\label{sec:main_results_ft}

% Although SHIFT-LLM is training-free, we further evaluate whether the proposed \gls{lra} remains beneficial when combined with post-pruning fine-tuning. We consider three settings: LoRA fine-tuning, partial-layer fine-tuning following the Navigation LLM protocol~\citep{lu2024reassessing}, and integration with the state-of-the-art Navigation LLM pipeline using the same pruning strategy, fine-tuning protocol, and hyperparameters for a fair comparison.
SHIFT-LLM itself requires no post-pruning fine-tuning; our primary setting is therefore the training-free evaluation in Sec.~\ref{sec:main_results_no_ft}. We additionally combine the \gls{lra} with existing fine-tuning schemes only to assess its compatibility with gradient-based recovery. We consider LoRA fine-tuning, partial-layer fine-tuning~\citep{lu2024reassessing}, and integration with the state-of-the-art Navigation LLM pipeline.

\paragraph{LoRA Fine-tuning.}
Table~\ref{tab:main_ft_reverse_lora} evaluates whether SHIFT-LLM remains complementary to LoRA-based post-pruning fine-tuning. Adding the \gls{lra} improves average zero-shot performance by \textbf{+1.41}, \textbf{+0.77}, \textbf{+1.14}, and \textbf{+0.72} points on Qwen2-1.5B, Qwen1.5-7B, Llama-3.1-8B-Instruct, and Vicuna-7B-v1.5, respectively. These results show that the explicit correction provided by the \gls{lra} remains complementary to parameter-efficient adaptation: the affine residual correction estimates the missing residual update, while fine-tuning further adapts the remaining model parameters.

\begin{table*}[t]
\centering
\caption{Zero-shot results with fine-tuning on C4 dataset under Reverse-order$^{*}$ pruning. ``Orig.'' denotes the unpruned model, ``Partial'' denotes standard post-pruning partial-layer fine-tuning, and ``Partial+LRA'' denotes partial-layer fine-tuning combined with the proposed \gls{lra}, whose affine residual correction is kept trainable during fine-tuning.}
\scriptsize
\setlength{\tabcolsep}{2.6pt}
\renewcommand{\arraystretch}{1.08}
% \begin{tabular}{lccc|ccc|ccc|ccc}
\begin{tabular}{l
c!{\color{gray!50}\vrule width 0.3pt}cc|
c!{\color{gray!50}\vrule width 0.3pt}cc|
c!{\color{gray!50}\vrule width 0.3pt}cc|
c!{\color{gray!50}\vrule width 0.3pt}cc}
\toprule
\multirow{2}{*}{Dataset}
& \multicolumn{3}{c|}{Qwen2-1.5B}
& \multicolumn{3}{c|}{Qwen1.5-7B}
& \multicolumn{3}{c|}{Llama-3.1-8B-It}
& \multicolumn{3}{c}{Vicuna-7B} \\
\cmidrule(lr){2-4} \cmidrule(lr){5-7} \cmidrule(lr){8-10} \cmidrule(lr){11-13}
& Orig. & Partial  & Partial +LRA
& Orig. & Partial  & Partial +LRA
& Orig. & Partial  & Partial +LRA
& Orig. & Partial  & Partial +LRA \\
\midrule
BoolQ      & 73.36 & \textbf{67.19} & 65.78 & 81.90 & \textbf{80.09} & 79.69 & 83.98 & \textbf{74.89} & 63.67 & 80.55 & \textbf{78.04} & 77.77 \\
PIQA       & 75.46 & 62.68 & \textbf{62.95} & 78.40 & 66.05 & \textbf{67.36} & 79.98 & 67.63 & \textbf{70.02} & 77.26 & \textbf{68.44} & 68.23 \\
HellaSwag  & 65.41 & 47.81 & \textbf{47.85} & 76.97 & 58.13 & \textbf{61.22} & 79.27 & 62.30 & \textbf{67.12} & 73.76 & 60.03 & \textbf{63.51} \\
WinoGrande & 66.22 & \textbf{54.70} & 53.99 & 66.14 & \textbf{58.72} & 57.38 & 74.19 & 60.22 & \textbf{64.01} & 69.61 & 60.14 & \textbf{62.19} \\
ARC-e      & 66.16 & \textbf{39.27} & 39.10 & 70.71 & 43.14 & \textbf{46.89} & 81.78 & 56.57 & \textbf{64.44} & 75.63 & 55.22 & \textbf{59.01} \\
ARC-c      & 36.09 & \textbf{28.33} & 28.07 & 42.83 & \textbf{34.73} & 34.39 & 55.03 & 39.16 & \textbf{44.80} & 45.82 & 36.09 & \textbf{39.25} \\
OBQA       & 36.40 & \textbf{28.80} & 27.00 & 41.80 & 31.60 & \textbf{32.40} & 43.00 & 35.60 & \textbf{39.20} & 45.00 & 35.40 & \textbf{37.20} \\
\midrule
Avg.       & 59.87 & \textbf{46.97} & 46.39 & 65.54 & 53.21 & \textbf{54.19} & 71.03 & 56.62 & \textbf{59.20} & 66.80 & 56.19 & \textbf{58.16} \\
\bottomrule
\end{tabular}
\label{tab:main_ft_reverse_partial}
\end{table*}

\begin{table*}[t]
\centering
\caption{Comparison with state-of-the-art pruning methods on Llama-3.1-8B-It using Alpaca dataset for fine-tuning. All previous results are taken from the Navigation LLM paper, and $^{*}$ indicates reproduced results reported in that work. ``Navigation LLM+LRA'' denotes Navigation LLM equipped with the proposed \gls{lra}.}
\scriptsize
\setlength{\tabcolsep}{2.7pt}
\renewcommand{\arraystretch}{1.14}
% \resizebox{\textwidth}{!}{%
\resizebox{0.9\textwidth}{!}{%
\begin{tabular}{l!{\color{gray!55}\vrule width 0.4pt}
                c!{\color{gray!55}\vrule width 0.4pt}
                c c c c c c c c c
                !{\color{gray!55}\vrule width 0.4pt}c}
\toprule
Method & PR & BoolQ & PIQA & HSwag & OBQA & ARC-e & ARC-c & MMLU & CMMLU & Wino & Avg. \\
\midrule
Original                  & 0\%  & 83.98    & 79.87 & 59.10 & 33.80 & 81.90 & 51.71 & 68.04 & 55.43 & 73.72 & 65.28 \\
ShortGPT (BI)             & 25\% & --    & 71.76 & 41.96 & 20.20 & 61.07 & 28.41 & 24.17 & 24.94 & 53.91 & 40.80 \\
Short. LLaMA (PPL)        & 25\% & --    & 76.28 & 49.31 & 26.40 & 72.90 & 38.05 & 33.67 & 27.24 & 57.93 & 47.72 \\
Short. LLaMA (Taylor)     & 25\% & --    & 71.38 & 49.64 & 27.40 & 68.48 & 41.81 & 28.61 & 25.04 & 71.35 & 47.96 \\
SLEB (FT)                 & 25\% & --    & 75.73 & 49.73 & 26.80 & 69.70 & 38.65 & 43.05 & 33.38 & 63.85 & 50.11 \\
LLM-Pruner$^{*}$          & 20\% & --    & 72.00 & 54.60 & --    & --    & --    & 25.30 & 25.00 & --    & 44.23 \\
SliceGPT$^{*}$            & 20\% & --    & 68.30 & 47.50 & --    & --    & --    & 28.80 & 24.80 & --    & 42.35 \\
LaCo$^{*}$                & 20\% & --    & 69.80 & 55.70 & --    & --    & --    & 26.50 & 25.20 & --    & 44.08 \\
LLM-Streamline$^{*}$      & 20\% & --    & 71.50 & 61.10 & -- & -- & -- & 45.50 & 29.40 & -- & 51.88 \\
GRASP$^{*}$               & 20\% & --    & 73.30 & 62.70 & -- & -- & -- & 43.10 & 30.70 & -- & 52.45 \\
\midrule
Navigation LLM* & 25\% & 67.61 & 73.67 &  \textbf{53.02} & \textbf{30.80} & 72.73 & \textbf{46.08} & \textbf{66.28} & \textbf{55.31} & 66.22 & 59.08 \\
Navigation LLM+\textbf{LRA} & 25\% & \textbf{69.21}  & \textbf{74.21} & 52.08 & 29.00 & \textbf{73.99} & 45.56 & 65.94 & 54.96 & \textbf{70.96} & \textbf{59.54} \\
\bottomrule
\end{tabular}%
}
\label{tab:llama_comparison}
\end{table*}

\paragraph{Partial-layer Fine-tuning.}
Table~\ref{tab:main_ft_reverse_partial} evaluates SHIFT-LLM with partial-layer fine-tuning under Reverse-order$^{*}$ pruning. Adding the \gls{lra} improves average zero-shot performance by \textbf{+0.98}, \textbf{+2.58}, and \textbf{+1.97} points on Qwen1.5-7B, Llama-3.1-8B-Instruct, and Vicuna-7B-v1.5, respectively, while Qwen2-1.5B shows a small decrease of $0.58$ points. 
Overall, the results show that SHIFT-LLM can provide additional recovery beyond partial-layer fine-tuning, with gains across most evaluated models.

\paragraph{Comparison with State-of-the-Art Pruning Methods.}
Navigation LLM~\citep{lu2024reassessing} combines Reverse-order pruning with partial-layer fine-tuning and provides a strong depth-pruning baseline. Keeping its pruning strategy, fine-tuning protocol, and hyperparameters unchanged, we only add the proposed \gls{lra}. 
As shown in Table~\ref{tab:llama_comparison}, this improves average accuracy from \textbf{59.08} to \textbf{59.54}, showing that adding the \gls{lra} does not degrade the strong fine-tuned baseline and can provide additional recovery.

\subsection{Ablation Studies}

We analyze three aspects of the proposed \gls{lra}: calibration set size, the importance of the residual parameterization, and robustness across pruning ratios. Additional ablations are provided in the supplementary material.

\begin{figure}[t]
    \centering
    \begin{subfigure}{0.5\columnwidth}
        \centering
        \includegraphics[width=\linewidth]{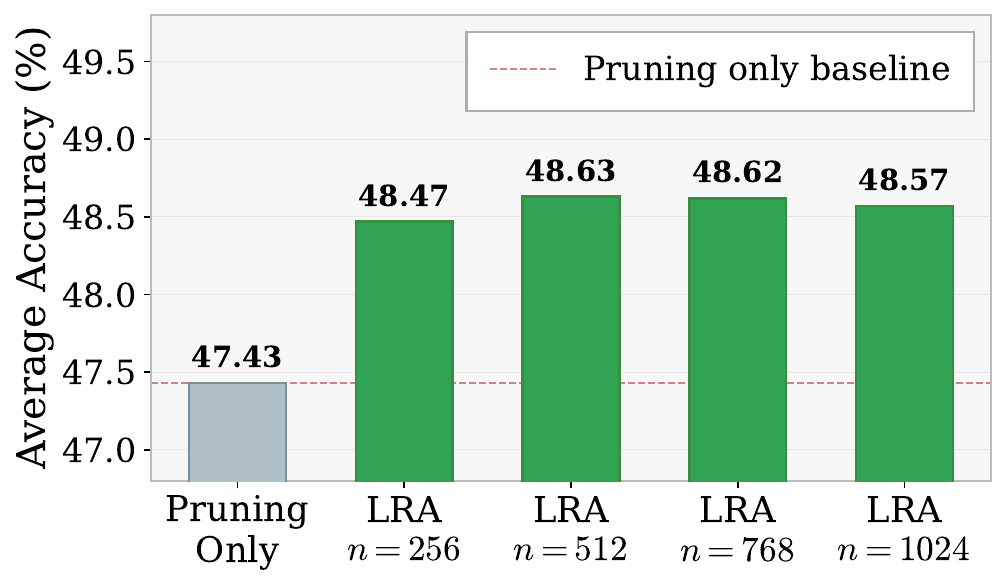}
        \caption{Effect of calibration set size}
        \label{fig:calibration_size}
    \end{subfigure}
    \hfill
    \begin{subfigure}{0.48\columnwidth}
        \centering
        \includegraphics[width=\linewidth]{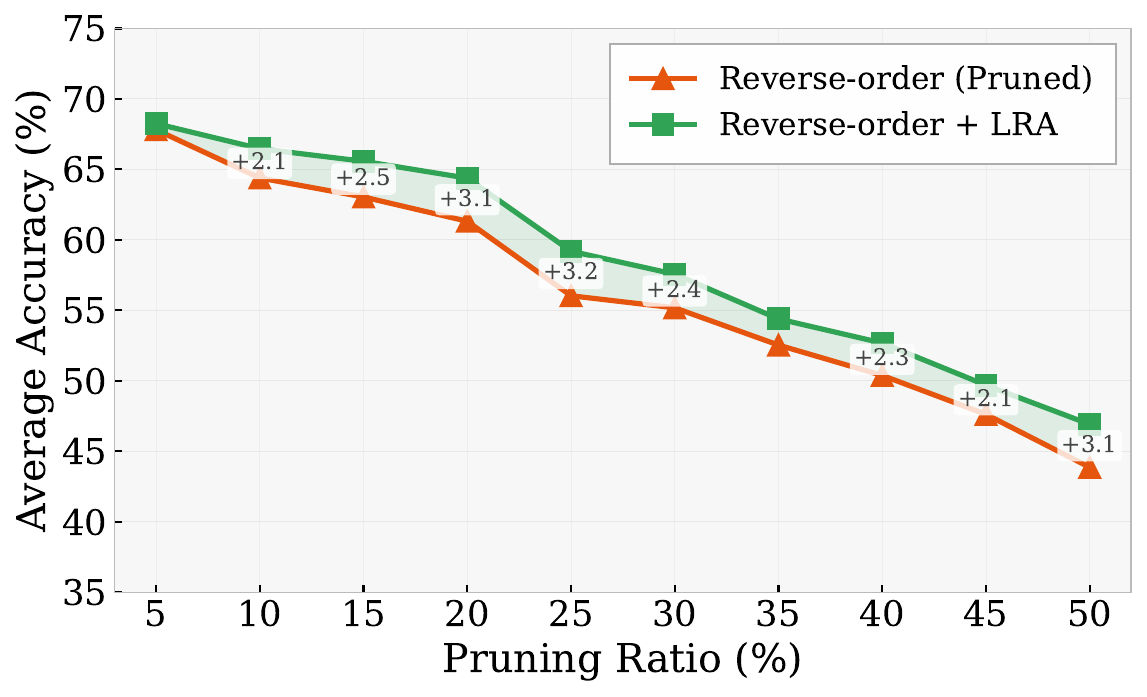}
        \caption{Effect of the pruning ratio}
        \label{fig:pruning_ratio}
    \end{subfigure}
    \caption{Ablation studies of the proposed \gls{lra} under different calibration sizes and pruning ratios.}
    \label{fig:pruning_ratio_all}
\end{figure}

\paragraph{Effect of Calibration Set Size.}
Figure~\ref{fig:calibration_size} evaluates calibration size under Magnitude-$\ell_1$ pruning on Qwen2-1.5B at a pruning ratio of 25\%. The proposed \gls{lra} improves over the pruning-only baseline for all calibration sizes, showing that its affine residual correction can be estimated effectively from limited calibration data. Since performance improves only slightly beyond 256 samples and quickly saturates, we use 256 calibration samples for all main experiments.

% \begin{wraptable}{r}{0.39\columnwidth}
%     \vspace{-10pt}
%     \scriptsize
%     \renewcommand{\arraystretch}{1.05}
%     \resizebox{\linewidth}{!}{%
%     \begin{tabular}{lccc}
%         \toprule
%         Criterion & Base & +LRA & +Generic \\
%         \midrule
%         BI & 45.78 & \textbf{46.85} & 36.96 \\
%         Reverse$^{*}$ & 43.71 & \textbf{45.71} & 37.99 \\
%         Magnitude-$\ell_1$ & 47.43 & \textbf{48.47} & 35.74 \\
%         \bottomrule
%     \end{tabular}%
%     }
%     \caption{LRA vs.\ Generic Affine on Qwen2-1.5B at 25\% pruning. Average zero-shot accuracy over seven benchmarks.}
%     \label{tab:generic_affine}
%     \vspace{-15pt}
% \end{wraptable}

\begin{wraptable}{r}{0.39\columnwidth}
    \vspace{-10pt}
    \scriptsize
    \renewcommand{\arraystretch}{1.05}
    \resizebox{\linewidth}{!}{%
    \begin{tabular}{lccc}
        \toprule
        Criterion & Base & +Generic  & +LRA \\
        \midrule
        BI & 45.78 & 36.96 &  \textbf{46.85} \\
        Reverse$^{*}$ & 43.71 & 37.99 &  \textbf{45.71}\\
        Magnitude-$\ell_1$ & 47.43 & 35.74 &  \textbf{48.47}\\
        \bottomrule
    \end{tabular}%
    }
    \caption{LRA vs.\ Generic Affine on Qwen2-1.5B at 25\% pruning. Average zero-shot accuracy over seven benchmarks.}
    \label{tab:generic_affine}
    \vspace{-15pt}
\end{wraptable}

% \paragraph{Importance of the Residual Parameterization.}
% We compare the proposed \gls{lra} with a Generic Affine baseline that directly fits
% $\hat{h}^{(\ell)}=h^{(\ell-1)}W^{(\ell)}+\mathbf{1}(b^{(\ell)})^\top$
% using the same calibration data. Although both formulations have equivalent affine representational capacity through $W^{(\ell)}=I+A^{(\ell)}$, the \gls{lra} preserves the identity pathway and estimates only the missing residual update. Table~\ref{tab:generic_affine} shows that this residual parameterization consistently performs better.

\paragraph{Importance of the Residual Parameterization.}
We compare the proposed \gls{lra} with a Generic Affine baseline that directly fits
$\hat{h}^{(\ell)}=h^{(\ell-1)}W^{(\ell)}+\mathbf{1}(b^{(\ell)})^\top$
using the same calibration data. Although both formulations have equivalent affine representational capacity through $W^{(\ell)}=I+A^{(\ell)}$, the \gls{lra} preserves the identity pathway and estimates only the missing residual update, providing a better residual inductive bias. Table~\ref{tab:generic_affine} shows that this parameterization consistently performs better.

\paragraph{Effect of the Pruning Ratio.}
Figure~\ref{fig:pruning_ratio} evaluates Reverse-order pruning on Llama-3.1-8B-Instruct across different pruning ratios. Although performance decreases as more layers are removed, SHIFT-LLM consistently outperforms the pruning-only baseline, typically by \textbf{+2} to \textbf{+3} average accuracy points.

\begin{wraptable}{r}{0.52\columnwidth}
    \vspace{-32pt}
    \centering
    \scriptsize
    \setlength{\tabcolsep}{2.5pt}
    \renewcommand{\arraystretch}{0.92}
    \resizebox{0.50\columnwidth}{!}{%
    \begin{tabular}{lcccc}
        \toprule
        \multicolumn{5}{c}{\textbf{Per-block computational cost}} \\
        \midrule
        Component & Params & FLOPs/token & Relative Cost & \\
        \midrule
        Transformer block  & 218M  & 436M & 100\%      & \\
        Full-rank LRA      & 16.8M & 33.6M & 7.7\%     & \\
        Rank-64 LRA        & 0.5M  & 1.0M  & $<0.25\%$ & \\
        \midrule
        \multicolumn{5}{c}{\textbf{End-to-end runtime at 25\% pruning}} \\
        \midrule
        Configuration & Params & Layers & Latency (ms) & Tokens/s \\
        \midrule
        Original             & 8.03B & 32 & 58.3 $\pm$ 0.3 & 21.4 $\pm$ 0.3 \\
        Pruned               & 6.29B & 24 & 44.7 $\pm$ 0.3 & 27.7 $\pm$ 0.3 \\
        Pruned + rank-64 LRA & 6.29B & 32 & 46.0 $\pm$ 2.4 & 26.1 $\pm$ 1.2 \\
        Pruned + merged LRA  & 6.30B & 25 & 44.9 $\pm$ 0.9 & 27.6 $\pm$ 0.3 \\
        \bottomrule
    \end{tabular}%
    }
    \caption{Computational efficiency of SHIFT-LLM on Llama-3.1-8B-Instruct.}
    \label{tab:efficiency}
    \vspace{-10pt}
\end{wraptable}

\subsection{Computational Efficiency}
\label{sec:computational_efficiency}

% As shown in Table~\ref{tab:efficiency}, a rank-64 \gls{lra}, used in all experiments reported in this paper, requires less than 0.25\% of the computation of a Llama-3.1-8B-Instruct block, with FLOPs excluding quadratic attention. Runtime is measured on a single V100 GPU. At 25\% depth pruning, the \gls{lra} preserves a \textbf{1.27$\times$} speedup over the original model, while exact merging reduces latency to 44.9\,ms, nearly matching the 44.7\,ms pruned-only model. Thus, SHIFT-LLM recovers accuracy with negligible additional inference cost.

As shown in Table~\ref{tab:efficiency}, a rank-64 \gls{lra}, used in all experiments reported in this paper, requires less than 0.25\% of the computation of a Llama-3.1-8B-Instruct block, with FLOPs excluding quadratic attention. Runtime is measured on a single V100 GPU. At 25\% depth pruning, the \gls{lra} preserves a \textbf{1.27$\times$} speedup over the original model while adding only a small runtime overhead to the pruned model. Exact merging further reduces latency to 44.9\,ms, nearly matching the 44.7\,ms pruned-only model. Thus, SHIFT-LLM provides accuracy recovery while largely preserving the inference-efficiency gains of depth pruning.

\section{Conclusion}

We presented SHIFT-LLM, a training-free framework for mitigating the hidden-state distribution shift caused by depth pruning. Its \acrfull{lra} preserves the identity pathway and uses a closed-form affine residual correction to approximate the missing residual update. Experiments across models, pruning criteria, and evaluation settings show consistent performance recovery while preserving the efficiency benefits of depth pruning.

\bibliographystyle{plain}
\bibliography{main}

% \section*{References}

% References follow the acknowledgments in the camera-ready paper. Use unnumbered first-level heading for
% the references. Any choice of citation style is acceptable as long as you are
% consistent. It is permissible to reduce the font size to \verb+small+ (9 point)
% when listing the references.
% Note that the Reference section does not count towards the page limit.
% \medskip

% {
% \small

% [1] Alexander, J.A.\ \& Mozer, M.C.\ (1995) Template-based algorithms for
% connectionist rule extraction. In G.\ Tesauro, D.S.\ Touretzky and T.K.\ Leen
% (eds.), {\it Advances in Neural Information Processing Systems 7},
% pp.\ 609--616. Cambridge, MA: MIT Press.

% [2] Bower, J.M.\ \& Beeman, D.\ (1995) {\it The Book of GENESIS: Exploring
%   Realistic Neural Models with the GEneral NEural SImulation System.}  New York:
% TELOS/Springer--Verlag.

% [3] Hasselmo, M.E., Schnell, E.\ \& Barkai, E.\ (1995) Dynamics of learning and
% recall at excitatory recurrent synapses and cholinergic modulation in rat
% hippocampal region CA3. {\it Journal of Neuroscience} {\bf 15}(7):5249-5262.
% }

%%%%%%%%%%%%%%%%%%%%%%%%%%%%%%%%%%%%%%%%%%%%%%%%%%%%%%%%%%%%
\newpage

\section*{\centering SHIFT-LLM \\ Distribution Shift Correction in Depth-Pruned LLMs \\ Supplementary Material}
\vspace{1cm}

\appendix

\section{Experimental Setup and Pruning Criteria}
\label{app:setup_pruning}

\subsection{Experimental Setup}

We evaluate SHIFT-LLM on five decoder-only LLMs: Qwen2-1.5B, Qwen1.5-7B, Qwen2.5-14B, Llama-3.1-8B-Instruct~\citep{grattafiori2024llama}, and Vicuna-7B-v1.5~\citep{zheng2023judging}. Unless otherwise stated, all SHIFT-LLM experiments use rank-64 \glspl{lra}. For data-driven pruning criteria (e.g., BI, Taylor, and PPL), we use 2,000 samples from the C4 validation split~\citep{dodge2021documenting}, while the affine residual corrections of the \glspl{lra} are fitted using 256 samples from the same split.

SHIFT-LLM itself requires no post-pruning fine-tuning. For the additional compatibility experiments with LoRA and partial-layer fine-tuning, we train on 10,000 samples from the C4 training split and validate on 200 samples from the C4 validation split, using a maximum sequence length of 512, a batch size of 25, a learning rate of $3{\times}10^{-5}$, and the standard causal language modeling objective. In partial-layer fine-tuning, the language modeling head and the last three unpruned layers are fine-tuned. For comparison with Navigation LLM~\citep{lu2024reassessing}, we follow its protocol and fine-tune on the full Alpaca-Cleaned dataset~\citep{taori2023alpaca} (${\sim}$52K instruction-response pairs) with a maximum sequence length of 256, a learning rate of $1{\times}10^{-5}$, an effective batch size of 64, and 2 training epochs.

For zero-shot evaluation, we report results on seven standard benchmarks: BoolQ~\cite{clark2019boolq}, PIQA~\cite{bisk2020piqa}, HellaSwag~\cite{zellers2019hellaswag}, WinoGrande~\cite{sakaguchi2021winogrande}, ARC-Easy~\cite{clark2018think}, ARC-Challenge~\cite{clark2018think}, and OpenBookQA~\cite{mihaylov2018can}. These tasks cover yes/no question answering, physical commonsense reasoning, sentence completion, coreference resolution, and science question answering. Following prior pruning work, we adopt a likelihood-based evaluation protocol. For BoolQ, we compare the log-likelihood of the continuations ``yes'' and ``no'' given the input context. For the remaining multiple-choice benchmarks, we compute the conditional log-probability of each candidate answer and select the highest-scoring option. For datasets with multi-token answer choices, namely HellaSwag, ARC-Challenge, and OpenBookQA, we additionally report length-normalized accuracy to reduce the bias toward shorter completions. All evaluations are performed on the full benchmark splits without subsampling. Importantly, none of the evaluation datasets is used for calibration or fine-tuning, ensuring a strict separation between adaptation and evaluation data. All experiments are conducted on NVIDIA V100 GPUs.
%Additional implementation details and hyperparameter settings are provided in the supplementary material.

\begin{table*}[t]
\centering
\caption{Per-benchmark zero-shot accuracy corresponding to the average results reported in Table~\ref{tab:main_no_ft_summary}. Results are shown for different pruning criteria at a pruning ratio of 25\%, without post-pruning fine-tuning. ``+LRA'' denotes the pruned model equipped with the proposed \gls{lra}.}
% \label{tab:per_benchmark_no_ft_all_models}
\scriptsize
\setlength{\tabcolsep}{3.2pt}
\renewcommand{\arraystretch}{0.82}
\resizebox{0.9\textwidth}{!}{%
\begin{tabular}{llc|ccccccc|c}
\toprule
Model & Method & PR & BoolQ & PIQA & HellaSwag & WinoGrande & ARC-e & ARC-c & OBQA & Avg. \\
\midrule
Qwen2-1.5B & Original & 0\% 
& 73.36 & 75.46 & 65.41 & 66.22 & 66.16 & 36.09 & 36.40 & 59.87 \\
\arrayrulecolor{gray!60}\cmidrule(lr){1-11}\arrayrulecolor{black}
Qwen1.5-7B & Original & 0\% 
& 81.90 & 78.40 & 76.97 & 66.14 & 70.71 & 42.83 & 41.80 & 65.54 \\
\arrayrulecolor{gray!60}\cmidrule(lr){1-11}\arrayrulecolor{black}
Llama-3.1-8B-It & Original & 0\% 
& 83.98 & 79.98 & 79.27 & 74.19 & 81.78 & 55.03 & 43.00 & 71.03 \\
\arrayrulecolor{gray!60}\cmidrule(lr){1-11}\arrayrulecolor{black}
Vicuna-7B & Original & 0\% 
& 80.55 & 77.26 & 73.76 & 69.61 & 75.63 & 45.82 & 45.00 & 66.80 \\
\midrule
\midrule
Qwen2-1.5B & ShortGPT (BI) & 25\%
& 55.57 & 62.51 & 43.17 & 55.01 & 45.20 & \textbf{28.58} & \textbf{30.40} & 45.78 \\
% \arrayrulecolor{gray!60}\cmidrule(lr){1-11}\arrayrulecolor{black}
Qwen2-1.5B & ShortGPT (BI)+LRA & 25\%
& \textbf{57.74} & \textbf{64.80} & \textbf{44.88} & \textbf{56.12} & \textbf{48.32} & 26.71 & 29.40 & \textbf{46.85} \\
\arrayrulecolor{gray!60}\cmidrule(lr){1-11}\arrayrulecolor{black}
Qwen1.5-7B & ShortGPT (BI) & 25\%
& 74.40 & 65.83 & \textbf{56.62} & 60.06 & 43.43 & \textbf{32.08} & \textbf{32.00} & 52.06 \\
% \arrayrulecolor{gray!60}\cmidrule(lr){1-11}\arrayrulecolor{black}
Qwen1.5-7B & ShortGPT (BI)+LRA & 25\%
& \textbf{76.54} & \textbf{67.03} & 55.54 & \textbf{63.93} & \textbf{45.50} & 32.00 & 29.20 & \textbf{52.82} \\
\arrayrulecolor{gray!60}\cmidrule(lr){1-11}\arrayrulecolor{black}
Llama-3.1-8B-It & ShortGPT (BI) & 25\%
& 77.74 & 68.23 & 60.62 & 66.46 & 54.80 & \textbf{40.78} & 31.60 & 57.18 \\
% \arrayrulecolor{gray!60}\cmidrule(lr){1-11}\arrayrulecolor{black}
Llama-3.1-8B-It & ShortGPT (BI)+LRA & 25\%
& \textbf{82.14} & \textbf{70.84} & \textbf{66.52} & \textbf{71.82} & \textbf{64.60} & 40.61 & \textbf{34.80} & \textbf{61.62} \\
\arrayrulecolor{gray!60}\cmidrule(lr){1-11}\arrayrulecolor{black}
Vicuna-7B & ShortGPT (BI) & 25\%
& 64.07 & 66.65 & \textbf{59.95} & 66.38 & 56.65 & \textbf{35.24} & \textbf{36.40} & 55.05 \\
% \arrayrulecolor{gray!60}\cmidrule(lr){1-11}\arrayrulecolor{black}
Vicuna-7B & ShortGPT (BI)+LRA & 25\%
& \textbf{69.63} & \textbf{66.92} & 56.63 & \textbf{66.85} & \textbf{58.63} & 34.81 & 35.20 & \textbf{55.52} \\
\midrule
\midrule
Qwen2-1.5B & Reverse-order$^{*}$ & 25\%
& 62.84 & 58.81 & 40.00 & 56.75 & 30.98 & 26.19 & \textbf{30.40} & 43.71 \\
% \arrayrulecolor{gray!60}\cmidrule(lr){1-11}\arrayrulecolor{black}
Qwen2-1.5B & Reverse-order$^{*}$+LRA & 25\%
& \textbf{64.77} & \textbf{62.08} & \textbf{41.25} & \textbf{59.98} & \textbf{36.20} & \textbf{26.88} & 28.80 & \textbf{45.71} \\
\arrayrulecolor{gray!60}\cmidrule(lr){1-11}\arrayrulecolor{black}
Qwen1.5-7B & Reverse-order$^{*}$ & 25\%
& 63.33 & 65.34 & 51.05 & 56.59 & 43.47 & 31.40 & \textbf{33.60} & 49.25 \\
% \arrayrulecolor{gray!60}\cmidrule(lr){1-11}\arrayrulecolor{black}
Qwen1.5-7B & Reverse-order$^{*}$+LRA & 25\%
& \textbf{79.27} & \textbf{67.52} & \textbf{53.57} & \textbf{63.30} & \textbf{47.26} & \textbf{33.62} & 32.00 & \textbf{53.79} \\
\arrayrulecolor{gray!60}\cmidrule(lr){1-11}\arrayrulecolor{black}
Llama-3.1-8B-It & Reverse-order$^{*}$ & 25\%
& 62.23 & 59.85 & 27.58 & 55.72 & 36.03 & 30.29 & 31.60 & 43.33 \\
% \arrayrulecolor{gray!60}\cmidrule(lr){1-11}\arrayrulecolor{black}
Llama-3.1-8B-It & Reverse-order$^{*}$+LRA & 25\%
& \textbf{62.42} & \textbf{70.84} & \textbf{64.26} & \textbf{71.82} & \textbf{64.23} & \textbf{44.11} & \textbf{35.80} & \textbf{59.07} \\
\arrayrulecolor{gray!60}\cmidrule(lr){1-11}\arrayrulecolor{black}
Vicuna-7B & Reverse-order$^{*}$ & 25\%
& 63.27 & \textbf{66.16} & \textbf{53.76} & 62.90 & \textbf{55.51} & \textbf{35.32} & \textbf{37.40} & \textbf{53.47} \\
% \arrayrulecolor{gray!60}\cmidrule(lr){1-11}\arrayrulecolor{black}
Vicuna-7B & Reverse-order$^{*}$+LRA & 25\%
& \textbf{68.56} & 64.36 & 51.86 & \textbf{64.33} & 53.32 & 34.47 & 33.80 & 52.96 \\
\midrule
\midrule
Qwen2-1.5B & Magnitude-$\ell_1$ & 25\%
& 61.74 & 65.83 & 43.02 & 51.93 & 51.89 & \textbf{26.45} & 31.20 & 47.44 \\
% \arrayrulecolor{gray!60}\cmidrule(lr){1-11}\arrayrulecolor{black}
Qwen2-1.5B & Magnitude-$\ell_1$+LRA & 25\%
& \textbf{62.29} & \textbf{67.68} & \textbf{45.37} & \textbf{53.99} & \textbf{52.57} & 25.60 & \textbf{31.80} & \textbf{48.47} \\
\arrayrulecolor{gray!60}\cmidrule(lr){1-11}\arrayrulecolor{black}
Qwen1.5-7B & Magnitude-$\ell_1$ & 25\%
& 56.02 & 54.57 & 34.77 & \textbf{51.70} & 35.14 & 26.02 & 26.80 & 40.72 \\
% \arrayrulecolor{gray!60}\cmidrule(lr){1-11}\arrayrulecolor{black}
Qwen1.5-7B & Magnitude-$\ell_1$+LRA & 25\%
& \textbf{61.59} & \textbf{66.16} & \textbf{40.39} & 51.54 & \textbf{55.56} & \textbf{26.11} & \textbf{31.80} & \textbf{47.59} \\
\arrayrulecolor{gray!60}\cmidrule(lr){1-11}\arrayrulecolor{black}
Llama-3.1-8B-It & Magnitude-$\ell_1$ & 25\%
& \textbf{54.83} & 55.43 & 27.09 & 48.62 & 25.17 & \textbf{25.51} & \textbf{28.00} & 37.81 \\
% \arrayrulecolor{gray!60}\cmidrule(lr){1-11}\arrayrulecolor{black}
Llama-3.1-8B-It & Magnitude-$\ell_1$+LRA & 25\%
& 49.82 & \textbf{59.58} & \textbf{30.10} & \textbf{51.70} & \textbf{37.04} & 22.10 & 25.60 & \textbf{39.42} \\
\arrayrulecolor{gray!60}\cmidrule(lr){1-11}\arrayrulecolor{black}
Vicuna-7B & Magnitude-$\ell_1$ & 25\%
& 37.83 & 52.34 & \textbf{26.43} & 47.99 & \textbf{26.01} & \textbf{27.05} & \textbf{26.40} & 34.86 \\
% \arrayrulecolor{gray!60}\cmidrule(lr){1-11}\arrayrulecolor{black}
Vicuna-7B & Magnitude-$\ell_1$+LRA & 25\%
& \textbf{39.05} & \textbf{53.05} & 26.14 & \textbf{49.96} & 25.34 & 26.88 & 25.20 & \textbf{35.09} \\
\midrule
\midrule
Qwen2-1.5B & Magnitude-$\ell_2$ & 25\%
& 61.47 & 62.73 & 37.04 & 50.99 & 43.39 & 23.29 & 27.40 & 43.76 \\
% \arrayrulecolor{gray!60}\cmidrule(lr){1-11}\arrayrulecolor{black}
Qwen2-1.5B & Magnitude-$\ell_2$+LRA & 25\%
& \textbf{62.05} & \textbf{64.58} & \textbf{41.04} & \textbf{51.70} & \textbf{45.79} & \textbf{23.89} & \textbf{31.40} & \textbf{45.78} \\
\arrayrulecolor{gray!60}\cmidrule(lr){1-11}\arrayrulecolor{black}
Qwen1.5-7B & Magnitude-$\ell_2$ & 25\%
& 60.52 & 60.83 & 36.69 & 50.51 & 44.49 & 24.57 & \textbf{29.20} & 43.83 \\
% \arrayrulecolor{gray!60}\cmidrule(lr){1-11}\arrayrulecolor{black}
Qwen1.5-7B & Magnitude-$\ell_2$+LRA & 25\%
& \textbf{61.96} & \textbf{64.36} & \textbf{36.90} & \textbf{52.17} & \textbf{49.33} & \textbf{26.88} & 28.60 & \textbf{45.74} \\
\arrayrulecolor{gray!60}\cmidrule(lr){1-11}\arrayrulecolor{black}
Llama-3.1-8B-It & Magnitude-$\ell_2$ & 25\%
& 45.57 & 51.80 & 27.08 & 49.72 & 26.05 & \textbf{25.00} & \textbf{26.80} & 36.00 \\
% \arrayrulecolor{gray!60}\cmidrule(lr){1-11}\arrayrulecolor{black}
Llama-3.1-8B-It & Magnitude-$\ell_2$+LRA & 25\%
& \textbf{53.00} & \textbf{59.09} & \textbf{29.02} & \textbf{51.78} & \textbf{37.63} & 21.16 & 25.40 & \textbf{39.58} \\
\arrayrulecolor{gray!60}\cmidrule(lr){1-11}\arrayrulecolor{black}
Vicuna-7B & Magnitude-$\ell_2$ & 25\%
& 37.92 & \textbf{52.77} & \textbf{26.04} & 48.86 & \textbf{25.13} & \textbf{28.41} & \textbf{26.60} & 35.10 \\
% \arrayrulecolor{gray!60}\cmidrule(lr){1-11}\arrayrulecolor{black}
Vicuna-7B & Magnitude-$\ell_2$+LRA & 25\%
& \textbf{54.74} & 52.45 & 25.80 & \textbf{50.36} & 24.96 & 26.28 & 24.60 & \textbf{37.03} \\
\midrule
\midrule
Qwen2-1.5B & Taylor & 25\%
& 62.72 & 62.40 & 42.02 & 57.85 & 38.76 & \textbf{27.99} & \textbf{29.80} & 45.93 \\
% \arrayrulecolor{gray!60}\cmidrule(lr){1-11}\arrayrulecolor{black}
Qwen2-1.5B & Taylor+LRA & 25\%
& \textbf{66.54} & \textbf{63.60} & \textbf{42.46} & \textbf{59.83} & \textbf{40.95} & 26.79 & 28.20 & \textbf{46.91} \\
\arrayrulecolor{gray!60}\cmidrule(lr){1-11}\arrayrulecolor{black}
Qwen1.5-7B & Taylor & 25\%
& 54.53 & 69.21 & \textbf{58.95} & 58.64 & 53.20 & \textbf{30.97} & 32.80 & 51.19 \\
% \arrayrulecolor{gray!60}\cmidrule(lr){1-11}\arrayrulecolor{black}
Qwen1.5-7B & Taylor+LRA & 25\%
& \textbf{62.35} & \textbf{70.40} & 57.33 & \textbf{59.27} & \textbf{57.53} & 30.80 & \textbf{33.60} & \textbf{53.04} \\
\arrayrulecolor{gray!60}\cmidrule(lr){1-11}\arrayrulecolor{black}
Llama-3.1-8B-It & Taylor & 25\%
& 62.23 & 59.74 & 27.56 & 55.80 & 35.98 & 30.20 & 31.60 & 43.30 \\
% \arrayrulecolor{gray!60}\cmidrule(lr){1-11}\arrayrulecolor{black}
Llama-3.1-8B-It & Taylor+LRA & 25\%
& \textbf{62.42} & \textbf{70.73} & \textbf{64.21} & \textbf{71.67} & \textbf{64.18} & \textbf{44.11} & \textbf{35.80} & \textbf{59.02} \\
\arrayrulecolor{gray!60}\cmidrule(lr){1-11}\arrayrulecolor{black}
Vicuna-7B & Taylor & 25\%
& 77.68 & \textbf{65.45} & \textbf{57.79} & 67.01 & \textbf{56.86} & \textbf{37.03} & 34.80 & 56.66 \\
% \arrayrulecolor{gray!60}\cmidrule(lr){1-11}\arrayrulecolor{black}
Vicuna-7B & Taylor+LRA & 25\%
& \textbf{79.82} & \textbf{65.45} & 57.70 & \textbf{67.40} & 56.69 & 34.64 & \textbf{35.00} & \textbf{56.67} \\
\midrule
\midrule
Qwen2-1.5B & PPL & 25\%
& 61.31 &	63.60 &	41.72 &	54.46 &	42.34 &	\textbf{27.90} &	28.8 & 45.73 \\
Qwen2-1.5B & PPL+LRA & 25\%
& \textbf{62.17} &	\textbf{65.23} &	\textbf{43.23} &	\textbf{58.72} &	\textbf{48.70} &	26.88 &	\textbf{29.40} & \textbf{47.76} \\
\arrayrulecolor{gray!60}\cmidrule(lr){1-11}\arrayrulecolor{black}
Qwen1.5-7B & PPL & 25\%
& 38.53	& 54.03	 & 27.31 &	49.33 &	26.39 &	\textbf{25.85} &	\textbf{29.8} & 35.89 \\
Qwen1.5-7B & PPL+LRA & 25\%
& \textbf{47.80} &	\textbf{56.75} &	\textbf{29.87} &	\textbf{50.75} &	\textbf{30.89} &	22.44 &	24.2 & \textbf{37.53} \\
\arrayrulecolor{gray!60}\cmidrule(lr){1-11}\arrayrulecolor{black}
Llama-3.1-8B-It & PPL & 25\%
& 80.09 &	72.36 &	65.56 &	70.17 &	63.17 &	\textbf{41.21} &	35.6 & 61.17 \\
Llama-3.1-8B-It & PPL+LRA & 25\%
& \textbf{81.71} &	\textbf{73.67} &	\textbf{66.66} &	\textbf{70.80} &	\textbf{65.99} &	40.87 &	\textbf{35.8} & \textbf{62.21} \\
\arrayrulecolor{gray!60}\cmidrule(lr){1-11}\arrayrulecolor{black}
Vicuna-7B & PPL & 25\%
& \textbf{75.38} &	\textbf{65.45} &	\textbf{57.48} &	\textbf{64.64} &	\textbf{55.01} &	\textbf{34.13} &	\textbf{33.80} & \textbf{55.13} \\
Vicuna-7B & PPL+LRA & 25\%
& 72.35 &	62.13 &	50.64 &	61.40 &	47.22 &	31.40 &	 30.40 & 50.79 \\
\bottomrule
\end{tabular}%
}
\label{tab:qwen15_full_results}
\end{table*}

\subsection{Layer Pruning Criteria}
\label{sec:pruning_criteria}

To evaluate the proposed method under diverse pruning settings, we consider several widely used criteria for selecting layers to remove. These criteria range from simple heuristics to data-dependent importance measures, allowing us to test the proposed \gls{lra}-based correction across a broad spectrum of pruning patterns.

Specifically, we consider six main layer selection strategies: \textit{Reverse-order}~\citep{men2025shortgpt}, \textit{Magnitude-$\ell_1$}~\citep{filters2016pruning, kim2024shortened, lu2024reassessing}, \textit{Magnitude-$\ell_2$}~\citep{filters2016pruning, kim2024shortened, lu2024reassessing}, \textit{Taylor}~\citep{lu2024reassessing}, \textit{Perplexity-based} (PPL)~\citep{lu2024reassessing}, and \textit{Block Influence} (BI)~\citep{men2025shortgpt}. The \textit{Reverse-order} strategy removes deeper layers first based on their position in the network. We also evaluate \textit{Reverse-order$^{*}$}, a modified variant that preserves the final Transformer layer, since its output feeds directly into the prediction head and cannot be corrected by a subsequent \gls{lra}. The two \textit{Magnitude}-based criteria rank layers according to the norms of their weights, using either the $\ell_1$ or $\ell_2$ norm. The \textit{Taylor} criterion estimates layer importance using a first-order approximation of the loss change on a calibration set. The \textit{PPL} criterion measures the increase in perplexity caused by removing a single layer and treats layers with smaller perplexity degradation as less important. Finally, \textit{Block Influence} measures how much a layer changes its input representation, using the average cosine similarity between the layer input and output as a proxy for redundancy.

Together, these criteria provide diverse pruning patterns for evaluating the robustness and generality of SHIFT-LLM across different layer-selection strategies.

\subsection{Per-Benchmark Results Without Fine-Tuning}
\label{app:per_benchmark_no_ft}

Table~\ref{tab:qwen15_full_results} provides the per-benchmark zero-shot results corresponding to the average scores reported in Table~\ref{tab:main_no_ft_summary}. Specifically, we report accuracy on BoolQ, PIQA, HellaSwag, WinoGrande, ARC-Easy, ARC-Challenge, and OpenBookQA for each pruning criterion at a pruning ratio of 25\%, without any post-pruning fine-tuning. The ``Original'' row denotes the unpruned model, while each pruning criterion is reported both before and after adding the proposed \gls{lra}. These detailed results show how the average improvements in the main paper are distributed across individual benchmarks.

\subsection{Results on Qwen2.5-14B}
\label{app:qwen14b}

To further evaluate scalability, we apply SHIFT-LLM to Qwen2.5-14B at 25\% depth pruning. Table~\ref{tab:qwen14b_full} reports the full per-benchmark results under Block Influence and Reverse-order$^{*}$ pruning. The proposed \gls{lra} improves average accuracy by \textbf{+2.98} points under BI and \textbf{+5.36} points under Reverse-order$^{*}$, showing that the training-free correction remains effective at the 14B scale.

\begin{table}[t]
\centering
\caption{Zero-shot results on Qwen2.5-14B at 25\% depth pruning.}
\label{tab:qwen14b_full}
\scriptsize
\setlength{\tabcolsep}{3.5pt}
\resizebox{0.85\columnwidth}{!}{%
\begin{tabular}{lcccccccc}
\toprule
Method & BoolQ & PIQA & HellaSwag & WinoGrande & ARC-e & ARC-c & OBQA & Avg. \\
\midrule
Original
& 86.27 & 81.07 & 82.91 & 75.14 & 82.45 & 58.87 & 45.20 & 73.13 \\
\midrule
BI
& 66.51 & 67.03 & 54.66 & 63.30 & 50.97 & 36.69 & \textbf{34.20} & 53.34 \\
BI + LRA
& \textbf{70.52} & \textbf{67.46} & \textbf{59.20} & \textbf{71.35} & \textbf{56.06} & \textbf{37.46} & 32.20 & \textbf{56.32} \\
\midrule
Reverse$^{*}$
& 62.23 & 59.19 & 42.48 & 58.01 & 40.32 & 31.83 & \textbf{33.00} & 46.72 \\
Reverse$^{*}$ + LRA
& \textbf{64.92} & \textbf{66.92} & \textbf{51.90} & \textbf{66.61} & \textbf{48.57} & \textbf{34.90} & 30.80 & \textbf{52.08} \\
\bottomrule
\end{tabular}%
}
\end{table}

\section{Additional Ablation Studies}
\label{app:additional_ablations}

\subsection{Effect of Low-Rank Factorization}
% Table~\ref{tab:low_rank_ablation_l1} examines the effect of applying low-rank factorization to the residual component of the proposed \gls{lra} under Magnitude-$\ell_1$ pruning on Qwen2-1.5B at a pruning ratio of 25\%, where performance is reported as the average zero-shot accuracy over seven evaluation datasets. Overall, the results show that this component can be compressed substantially without degrading performance. In fact, all low-rank variants slightly outperform the full-rank version in terms of average accuracy, with the best result achieved at rank 128. This behavior suggests that the fitted correction matrix contains redundant or noisy directions that can be removed without harming its ability to compensate for pruning-induced hidden-state shift. Based on this efficiency--performance trade-off, we use rank 64 in all main experiments.
Table~\ref{tab:low_rank_ablation_l1} examines the effect of low-rank factorization on the affine residual correction of the proposed \gls{lra} under Magnitude-$\ell_1$ pruning on Qwen2-1.5B at 25\% depth pruning. Performance is reported as the average zero-shot accuracy over seven benchmarks. The low-rank variants match or slightly outperform the full-rank \gls{lra}, suggesting that the fitted correction contains directions that can be removed without degrading its ability to approximate the missing residual update. Rank 128 achieves the highest accuracy, while rank 64 provides a stronger efficiency--performance trade-off and is therefore used in all reported experiments unless otherwise stated.

\begin{table*}[t]
\centering
\caption{Effect of low-rank factorization of the affine residual correction under Magnitude-$\ell_1$ pruning on Qwen2-1.5B at 25\% depth pruning.}
\footnotesize
\setlength{\tabcolsep}{4pt}
\renewcommand{\arraystretch}{1.1}
\begin{tabular}{lcccccccc}
\toprule
LRA Configuration & BoolQ & PIQA & HellaSwag & WinoGrande & ARC-e & ARC-c & OBQA & Avg. \\
\midrule
Full-rank & 62.23 & 67.95 & 43.25& 54.38 & 52.99 & 25.60 & 32.20 & 48.38 \\
\rowcolor{blue!20} Rank 64          & 62.29 & 67.68 & 45.37 & 53.99 & 52.57 & 25.60 & 31.80 & 48.47 \\
Rank 128         & 62.26 & 68.44 & 45.65 & 52.96 & 53.66 & 25.51 & 32.20 & \textbf{48.67} \\
Rank 256         & 62.26 & 69.04 & 45.18 & 52.57 & 53.79 & 25.51 & 31.80 & 48.59 \\
Rank 512         & 62.26 & 68.39 & 44.21 & 53.59 & 53.87 & 25.34 & 32.20 & 48.55 \\
\bottomrule
\end{tabular}
\label{tab:low_rank_ablation_l1}
\end{table*}

\begin{table}[b]
\centering
\caption{Sensitivity to the calibration domain on Vicuna-7B at 25\% depth pruning. Results are average zero-shot accuracy over seven benchmarks.}
\label{tab:calibration_domain}
\small
\begin{tabular}{lcccc}
\toprule
Criterion & Pruned & +LRA (C4) & +LRA (WikiText) & +LRA (Alpaca) \\
\midrule
BI & 55.05 & 55.66 & {56.14} & 54.66 \\
Reverse$^{*}$ & 53.47 & 52.96 & {53.36} & 52.20 \\
Magnitude-$\ell_1$ & 34.86 & 35.09 & {35.80} & 35.17 \\
Magnitude-$\ell_2$ & 35.10 & 37.02 & {38.16} & 37.32 \\
\bottomrule
\end{tabular}
\end{table}

\subsection{Calibration-Domain Sensitivity}
\label{app:calibration_domain}

To determine whether the weaker results on Vicuna-7B are caused by a mismatch between the C4 calibration data and the model's instruction-tuning distribution, we fit the \gls{lra} using C4, WikiText, and Alpaca calibration samples. Table~\ref{tab:calibration_domain} shows that changing the calibration domain does not eliminate the negative gain under Reverse-order$^{*}$ pruning, and differences across calibration sources remain relatively small in most settings. In particular, instruction-aligned Alpaca calibration does not consistently outperform C4. These results indicate that the weaker Vicuna results are not primarily caused by calibration-domain mismatch.

\subsection{Random Layer Selection and Failure Mode}
\label{app:random_pruning}

We further examine when the token-wise affine correction fails by applying the \gls{lra} after random layer removal. Unlike principled pruning criteria, random selection may remove blocks whose missing residual updates contain stronger nonlinear or cross-token structure. As shown in Table~\ref{tab:random_pruning}, the \gls{lra} decreases average accuracy from 38.22 to 36.07 in this setting. This result supports the interpretation that SHIFT-LLM is most effective when the pruning criterion selects blocks whose missing residual updates are sufficiently amenable to low-complexity approximation.

\begin{table}[t]
\centering
\caption{Random pruning with and without the proposed \gls{lra} on Qwen2-1.5B at 25\% depth pruning.}
\label{tab:random_pruning}
\small
\begin{tabular}{lcccccccc}
\toprule
Method & BoolQ & PIQA & HSwag & Wino & ARC-e & ARC-c & OBQA & Avg. \\
\midrule
Random
& 57.83 & 55.82 & 27.42 & 50.59 & 28.41 & 22.87 & 24.60 & 38.22 \\
Random + LRA
& 38.84 & 54.84 & 27.60 & 48.70 & 31.06 & 22.87 & 28.40 & 36.07 \\
\bottomrule
\end{tabular}
\end{table}

\subsection{Independent vs.\ Sequential Calibration}
\label{app:sequential_calibration}

% \begin{wraptable}{r}{0.35\columnwidth}
%     \vspace{-10pt}
%     \centering
%     \scriptsize
%     \setlength{\tabcolsep}{4pt}
%     \renewcommand{\arraystretch}{1.05}
%     \resizebox{\linewidth}{!}{%
%     \begin{tabular}{llc}
%         \toprule
%         Criterion & Calibration & Avg. \\
%         \midrule
%         BI & Independent & \textbf{56.62} \\
%         BI & Sequential & 56.29 \\
%         Reverse$^{*}$ & Independent & 54.61 \\
%         Reverse$^{*}$ & Sequential & \textbf{55.03} \\
%         \bottomrule
%     \end{tabular}%
%     }
%     \caption{Independent vs.\ sequential LRA calibration on Llama-3.1-8B-Instruct at 25\% depth pruning.}
%     \label{tab:sequential_calibration}
%     \vspace{-25pt}
% \end{wraptable}

\begin{table}[t]
\centering
\caption{Independent vs.\ sequential LRA calibration on Llama-3.1-8B-Instruct at 25\% depth pruning.}
\label{tab:sequential_calibration}
\scriptsize
\setlength{\tabcolsep}{3.5pt}
\renewcommand{\arraystretch}{1.05}
\resizebox{0.99\columnwidth}{!}{%
\begin{tabular}{llcccccccc}
\toprule
Criterion & Calibration & BoolQ & PIQA & HellaSwag & WinoGrande & ARC-e & ARC-c & OBQA & Avg. \\
\midrule
BI & Independent
& 82.14 & 70.84 & 46.21 & 71.82 & 64.60 & 38.14 & 22.60 & \textbf{56.62} \\

BI & Sequential
& 83.21 & 70.13 & 45.53 & 72.30 & 64.06 & 37.03 & 21.80 & 56.29 \\
\midrule

Reverse$^{*}$ & Independent
& 62.42 & 70.84 & 46.12 & 71.82 & 64.23 & 41.21 & 25.60 & 54.61 \\

Reverse$^{*}$ & Sequential
& 62.54 & 71.87 & 46.15 & 72.61 & 66.62 & 41.21 & 24.20 & \textbf{55.03} \\
\bottomrule
\end{tabular}%
}
\end{table}

Our default procedure estimates each \gls{lra} independently from activations of the original model. To test whether accumulated corrections create a calibration--inference mismatch, we also perform sequential calibration, where each fitted \gls{lra} is inserted before collecting activations for the next pruning site. Table~\ref{tab:sequential_calibration} shows nearly identical performance between the two strategies. This indicates that the corrected hidden states remain sufficiently close to the original representations for independent closed-form calibration to be effective.

\subsection{Robustness to Calibration Samples}
\label{app:calibration_seeds}

Because the \gls{lra} is estimated in closed form, identical calibration data produces identical parameters and no optimization randomness is involved. We therefore test sensitivity to the composition of the calibration set using two random calibration seeds on Llama-3.1-8B-Instruct. The resulting gains remain consistent: BI improves by +4.44 and +5.09 points, while Reverse-order$^{*}$ improves by +15.74 and +16.21 points. The variation remains below one point, indicating that the training-free correction is robust to the sampled calibration examples.

\begin{table}[b]
\centering
\caption{Frozen vs.\ trainable \glspl{lra} during LoRA fine-tuning under Reverse-order$^{*}$ pruning at 25\% depth pruning.}
\label{tab:train_vs_freeze}
\small
\begin{tabular}{lccc}
\toprule
Model & Frozen & Trainable & Gain \\
\midrule
Qwen2-1.5B & 47.98 & \textbf{48.94} & +0.96 \\
Qwen1.5-7B & 56.08 & \textbf{56.28} & +0.20 \\
Llama-3.1-8B-It & 60.32 & \textbf{61.12} & +0.80 \\
\bottomrule
\end{tabular}
\end{table}

\subsection{Training vs.\ Freezing the LRAs During Fine-Tuning}
\label{app:frozen_trainable}

Although SHIFT-LLM itself is training-free, we additionally examine whether its closed-form solution can serve as a useful initialization when post-pruning fine-tuning is available. Table~\ref{tab:train_vs_freeze} compares freezing the fitted \glspl{lra} with allowing their affine residual corrections to remain trainable during LoRA fine-tuning. Training the correction consistently provides additional gains, indicating that the closed-form solution can be further refined when gradient-based recovery is desired.

% \subsection{Effect of the Reverse Pruning Variant}
% We also compare two variants of Reverse pruning on Qwen2-1.5B, denoted as Reverse-order and Reverse-order$^{*}$. Here, Reverse-order corresponds to the original strategy, in which the deepest layers are removed directly, while Reverse-order$^{*}$ denotes our modified version, where the final Transformer layer is preserved and the preceding deep layers are pruned instead. The difference is substantial: under Reverse-order$^{*}$, the proposed \gls{lra} improves the average accuracy from \textbf{43.71} to \textbf{45.71} (\textbf{+2.00}), whereas under the original Reverse-order strategy it decreases slightly from \textbf{44.36} to \textbf{43.80} (\textbf{-0.56}). Although the pruning-only baseline is marginally lower for Reverse-order$^{*}$, the final corrected model performs better overall. A plausible explanation is that pruning the last layer disrupts the final alignment between deep hidden states and the language modeling head, making the resulting shift harder to approximate with a lightweight residual correction. By preserving the final layer, Reverse-order$^{*}$ retains this last alignment stage, so the \gls{lra} only needs to correct an earlier and more structured hidden-state shift, which is more compatible with the proposed residual formulation.

\subsection{Effect of the Reverse Pruning Variant}
\label{app:reverse_variant}

We compare the original Reverse-order strategy with Reverse-order$^{*}$ on Qwen2-1.5B. Reverse-order removes the deepest layers directly, whereas Reverse-order$^{*}$ preserves the final Transformer block and instead removes the preceding deep layers. Under Reverse-order$^{*}$, the proposed \gls{lra} improves average accuracy from 43.71 to 45.71 (\textbf{+2.00}), whereas under the original Reverse-order strategy the accuracy decreases slightly from 44.36 to 43.80 ($-0.56$). Preserving the final block retains the pretrained transformation immediately preceding the language-modeling head, while the \gls{lra} corrects earlier missing residual updates. We therefore use Reverse-order$^{*}$ in the main experiments.

\begin{table}[t]
\centering
\caption{Comparison with ReplaceMe (L2) at 25\% depth pruning. Results are average zero-shot accuracy over seven benchmarks using the same 256-sample C4 calibration budget.}
\label{tab:replaceme_all_models}
\small
\setlength{\tabcolsep}{4pt}
\begin{tabular}{llccc}
\toprule
Model & Criterion & Pruned & ReplaceMe & SHIFT-LLM \\
\midrule
Qwen2-1.5B
& BI
& 45.8
& 43.9 {\scriptsize $(-1.9)$}
& \textbf{46.9} {\scriptsize $(+1.1)$} \\

Qwen2-1.5B
& Reverse$^{*}$
& 43.7
& 44.3 {\scriptsize $(+0.6)$}
& \textbf{45.7} {\scriptsize $(+2.0)$} \\
\midrule

Qwen1.5-7B
& BI
& 52.06
& 50.34 {\scriptsize $(-1.72)$}
& \textbf{52.82} {\scriptsize $(+0.76)$} \\

Qwen1.5-7B
& Reverse$^{*}$
& 49.25
& 53.62 {\scriptsize $(+4.37)$}
& \textbf{53.79} {\scriptsize $(+4.54)$} \\
\midrule

Llama-3.1-8B
& BI
& 57.18
& 59.46 {\scriptsize $(+2.28)$}
& \textbf{61.62} {\scriptsize $(+4.44)$} \\

Llama-3.1-8B
& Reverse$^{*}$
& 43.33
& 58.02 {\scriptsize $(+14.69)$}
& \textbf{59.07} {\scriptsize $(+15.74)$} \\
\midrule

Vicuna-7B
& BI
& 55.05
& 54.33 {\scriptsize $(-0.72)$}
& \textbf{55.52} {\scriptsize $(+0.48)$} \\

Vicuna-7B
& Reverse$^{*}$
& 53.47
& 46.40 {\scriptsize $(-7.07)$}
& \textbf{52.95} {\scriptsize $(-0.52)$} \\
\bottomrule
\end{tabular}
\end{table}

\subsection{Comparison with a Closely Related Linear Recovery Baseline}
\label{sec:replaceme_comparison}

Among training-free depth-pruning methods, ReplaceMe~\citep{shopkhoev2025replaceme} is closely related to SHIFT-LLM: both use a small calibration set to estimate a lightweight linear correction after block removal, without gradient-based training. However, the two methods differ in both the representation used for regression and the quantity directly approximated.

ReplaceMe considers a retained Transformer block $i$, whose post-attention representation and MLP contribution are denoted by $Y_i$ and $M_i$, respectively. After removing the subsequent blocks $i+1,\ldots,i+n$, it estimates a linear transformation $T$ by solving
\begin{equation}
    T^{*}
    =
    \arg\min_T
    \left\|
        M_iT + Y_i - L_{i+n}
    \right\|_F^2,
    \label{eq:replaceme}
\end{equation}
where $L_{i+n}$ is the original hidden state after the removed block sequence. Thus, ReplaceMe transforms the MLP contribution $M_i$ of the retained predecessor block, while $Y_i$ remains a fixed additive term, so that their sum directly approximates the original output after the removed span. This matches the formulation in ReplaceMe, where the learned transformation is applied to the predecessor MLP output and the blocks $i+1,\ldots,i+n$ are bypassed. 

SHIFT-LLM instead performs a local correction at each pruning site. For a removed block $\ell$, the original Transformer block computes
\begin{equation}
    h_{\mathrm{out}}^{(\ell)}
    =
    h_{\mathrm{in}}^{(\ell)}
    +
    \Delta_\ell,
    \qquad
    \Delta_\ell
    =
    f_\ell\!\left(h_{\mathrm{in}}^{(\ell)}\right),
\end{equation}
where $\Delta_\ell$ denotes the missing residual update introduced by removing block $\ell$. Rather than directly regressing the full output hidden state, SHIFT-LLM estimates only this missing residual update using the full hidden state at the pruning site:
\begin{equation}
    (A^{*},b^{*})
    =
    \arg\min_{A,b}
    \left\|
        h_{\mathrm{in}}^{(\ell)}A
        +
        \mathbf{1}b^\top
        -
        \Delta_\ell
    \right\|_F^2
    +
    \lambda\|A\|_F^2.
    \label{eq:ours_fit}
\end{equation}
The resulting \gls{lra} is
\begin{equation}
    \hat h_{\mathrm{out}}^{(\ell)}
    =
    h_{\mathrm{in}}^{(\ell)}
    +
    h_{\mathrm{in}}^{(\ell)}A^{*}
    +
    \mathbf{1}(b^{*})^\top.
\end{equation}

The two methods therefore differ in two important ways. First, ReplaceMe learns its transformation from the MLP contribution $M_i$ of the retained predecessor block, whereas SHIFT-LLM uses the full hidden state $h_{\mathrm{in}}^{(\ell)}$ at the pruning site as the input to its affine residual correction. Second, ReplaceMe directly fits the full hidden state $L_{i+n}$ after the removed span through $Y_i+M_iT$, whereas SHIFT-LLM explicitly preserves $h_{\mathrm{in}}^{(\ell)}$ as the identity pathway and estimates only the missing residual update $\Delta_\ell$. The affine bias $b^{*}$ additionally allows SHIFT-LLM to capture a constant offset in the missing residual update.

These design choices provide a more direct local approximation of each removed block's missing residual update: the correction is conditioned on the full hidden state that originally drives $f_\ell$, while the identity pathway is preserved explicitly and only the missing residual update is estimated.

The importance of the identity-plus-correction parameterization is also supported by the Generic Affine ablation in Table~\ref{tab:generic_affine}. Although a generic affine mapping has the same representational capacity as the proposed parameterization, it performs substantially worse in our experiments. This suggests that explicitly preserving the identity pathway and estimating only the missing residual update provides a beneficial parameterization for post-pruning recovery.

Table~\ref{tab:replaceme_all_models} provides a direct comparison using the same 256-sample C4 calibration budget. Across all evaluated model--pruning combinations, SHIFT-LLM achieves higher average zero-shot accuracy than ReplaceMe (L2). The difference is particularly large on Vicuna-7B under Reverse-order$^{*}$ pruning, where ReplaceMe substantially degrades the pruning-only model, while SHIFT-LLM produces a considerably smaller degradation.

\section{Additional Computational Analysis}
\label{app:computational_cost}

As reported in Table~\ref{tab:efficiency} of the main paper, SHIFT-LLM introduces only a small computational overhead relative to the removed Transformer blocks. Here, we provide additional details on how this overhead scales with the number of pruned layers.

For a Llama-3.1-8B-Instruct block, the rank-64 \gls{lra} requires approximately 0.5M parameters and 1.0M FLOPs per token, compared with 218M parameters and 436M FLOPs per token for the original Transformer block, excluding the quadratic attention term. More generally, for a pruning ratio $\rho$, depth pruning removes $\lfloor\rho L\rfloor$ Transformer blocks, while the \gls{lra} overhead grows only linearly with the number of pruning sites.

At 25\% pruning of the 32-layer Llama-3.1-8B-Instruct model, eight Transformer blocks are removed, corresponding to approximately 1.74B parameters and 3.49G FLOPs per token. Replacing the eight pruning sites with rank-64 \glspl{lra} adds back only approximately 4.2M parameters and 8.4M FLOPs per token, i.e., less than 0.25\% of the removed computational cost. When multiple pruned layers are consecutive, their affine mappings can additionally be merged exactly into a single transformation, as described in the main paper.

\subsection{Runtime Measurement Details}

Runtime measurements reported in Table~\ref{tab:efficiency} of the main paper are obtained on a single NVIDIA V100 GPU using an input length of 128 tokens and greedy generation of 128 output tokens. The reported latency and throughput values are measured for the original model, the depth-pruned model, the pruned model with rank-64 \glspl{lra}, and the exactly merged configuration.
%%%%%%%%%%%%%%%%%%%%%%%%%%%%%%%%%%%%%%%%%%%%%%%%%%%%%%%%%%%%

% \newpage
% \input{checklist.tex}

\end{document}